\documentclass{article}
\pdfoutput=1 

\PassOptionsToPackage{numbers, sort&compress, square, comma}{natbib} 
\usepackage[final]{neurips_2020}

\usepackage[utf8]{inputenc} 
\usepackage[T1]{fontenc}    
\usepackage{hyperref}       
\hypersetup{
    colorlinks=true,
    linkcolor=black, 
    filecolor=magenta,      
    urlcolor=blue,
    citecolor=black
}

\usepackage{url}            
\usepackage{booktabs}       
\usepackage{tabularx}       
\usepackage{multirow}
\usepackage{amsfonts}       
\usepackage{nicefrac}       
\usepackage{microtype}      
\usepackage{epigraph}
\usepackage{soul}
\usepackage[dvipsnames]{xcolor}
\usepackage{paralist}
\usepackage{graphicx}
\usepackage{subcaption}
\usepackage{color, colortbl}
\usepackage{amsmath}
\usepackage{amssymb}
\usepackage{pifont} 
\usepackage{wrapfig}

\usepackage{makecell}
\usepackage{array}
\usepackage[export]{adjustbox}

\newcommand{\cit}[1]{\cite{#1}}

\newcommand{\coot}{COOT}

\definecolor{mygreen}{rgb}{0.032, 0.6392, 0.2039}
\newcommand{\cmark}{\textcolor{mygreen}{\ding{51}}}%
\newcommand{\xmark}{\textcolor{red}{\ding{55}}}%

\newcommand{\narrowparagraph}[1]{\vspace{-2mm}\paragraph{#1}}

\newcommand{\std}[1]{\tiny{$\pm$#1}}
\newcommand{\nsn}[2]{#1\std{#2}} 
\newcommand{\nsb}[2]{\textbf{#1}\std{#2}} 

\newcommand{\tbf}[1]{\textbf{#1}}

\newcommand{\mr}[1]{\multirow{2}{*}{#1}}

\newcommand{\msb}[2]{\multirow{2}{*}{\nsb{#1}{#2}}}

\newcommand{\markres}[1]{\textbf{\color{ForestGreen}{#1}}}
\newcommand{\imagequal}[1]{\includegraphics[width=\linewidth,valign=t]{retrieval_supp/#1}}
\newcommand{\imagequeryqual}[1]{\includegraphics[width=\dimexpr 0.5\textwidth-\tabcolsep,valign=t]{retrieval_supp/#1}}

\newcommand{\multicolqual}[2]{\multicolumn{2}{ >{\hsize=\dimexpr 0.5\textwidth-2\tabcolsep} #1}{#2}}
\newcommand{\rankscoretitle}[2]{\makecell[tc]{#1\\{\textit{#2}}}}
\newcommand{\rankscore}[2]{\rankscoretitle{\textbf{#1}}{#2}}
\newcommand{\rankscoremark}[2]{\rankscoretitle{\markres{#1}}{#2}}

\newcommand{\arraystretchqual}{1.3}
\newcommand{\arraystretchquant}{1.2}
\newcommand{\tabcolsepqual}{0.1cm}
\newcommand{\tabcolsepquant}{0.1cm}

\newcommand{\thickhline}{
    \specialrule{.1em}{0em}{0em}
}

\newcommand{\ignore}[1]{} 

\graphicspath{ {./images/} }

\title{\coot: Cooperative Hierarchical Transformer for Video-Text
Representation Learning}

\author{%
    Simon Ging$^1$\textsuperscript{*}, Mohammadreza Zolfaghari$^1$\textsuperscript{*},  Hamed Pirsiavash$^2$, Thomas Brox$^1$ \\
    $^1$University of Freiburg, $^2$University of Maryland Baltimore County \\
    \texttt{$^1$\{gings, zolfagha, brox\}@cs.uni-freiburg.de}, $^2$ \texttt{hpirsiav@umbc.edu} \\
}

\begin{document}

    \maketitle

    \begin{abstract}
    Many real-world video-text tasks involve different levels of granularity, such as frames and words, clip and
    sentences or videos and paragraphs, each with distinct semantics.
    In this paper, we propose a \textbf{Coo}perative hierarchical \textbf{T}ransformer
    (\textbf{\coot}) to leverage this hierarchy information and model the interactions between different levels
    of granularity and different modalities. The method consists of three major components:
    an attention-aware feature aggregation layer, which leverages the local temporal context
    (intra-level, e.g., within a clip),
    a contextual transformer to learn the interactions between low-level and high-level semantics
    (inter-level, e.g. clip-video, sentence-paragraph),
    and a cross-modal cycle-consistency loss to connect video and text.
    The resulting method compares favorably to the state of the art on several benchmarks while having few parameters.
    All code is available open-source at \url{https://github.com/gingsi/coot-videotext}
\end{abstract}

    \section{Introduction}\label{sec:introduction}
{
\renewcommand{\thefootnote}{\fnsymbol{footnote}}
\footnotetext[1]{Equal contribution}
}
Representation learning based on both vision and language has many potential benefits: visual
grounding\cite{KotturVMP15,chenlook20,rohrbach2016grounding,gad13}; visual learning with a more natural, almost
self-supervised annotation process; and direct applicability to cross-modal tasks, such as video retrieval by
text\cite{Shao_2018_ECCV,chen2020finegrained,zhang2018man,actbert20,videotextretriv_1}, video
summarization~\cite{8099601}, and automated indexing. This research direction has recently
boomed~\cite{sun2019videobert,li2019visualbert,lu2019vilbert,tan2019lxmert,su2019vl,actbert20,miech19howto100m,
miech20endtoend} also due to the success of self-attention in text analysis~\cite{attention_nips17, devlin2018bert}
with its almost immediate applicability in the cross-modal context. Many different research foci are currently
developing in this area, where some are concerned with large-scale pretraining to leverage the abundant data
available~\cite{miech20endtoend,actbert20,miech19howto100m,sun2019videobert} to learn a joint embedding space, and
others to bring in more explicit structure~\cite{Liu2019a,cmhse,hierarchical_3} or new losses~\cite{miech20endtoend,
sun2019contrastive} into the learning process.

\begin{figure*}[t]
    \centering
    \includegraphics[width=1.0\textwidth]{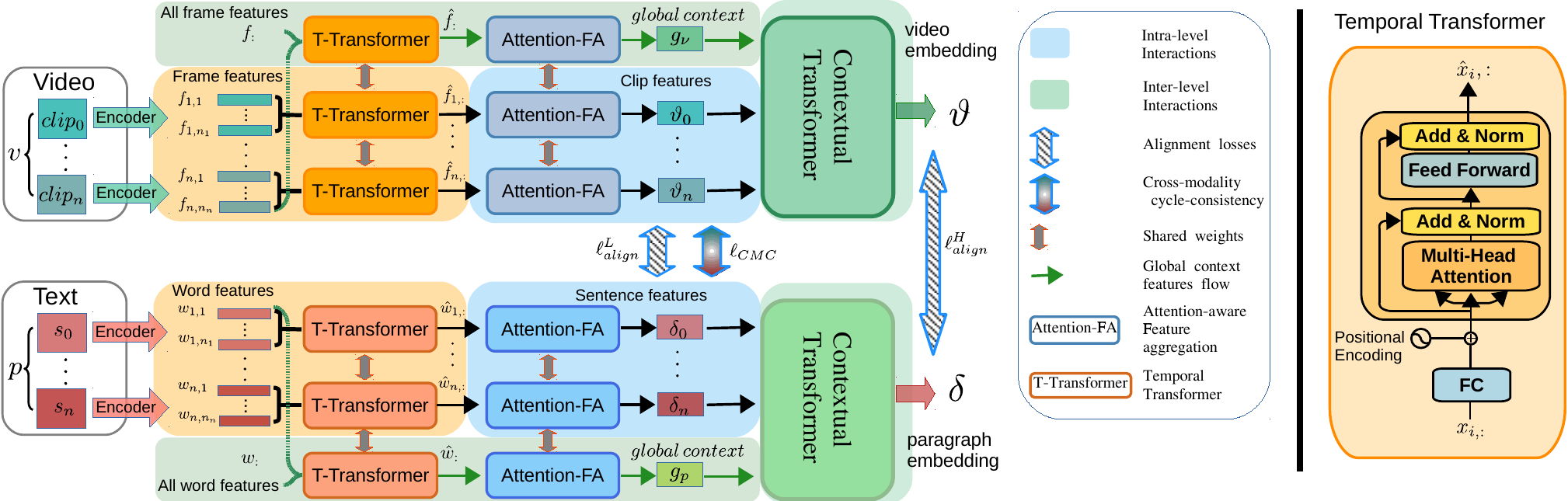}

    \caption{
    \textbf{Overview of \coot\ model} (best viewed in color). The model consist of two branches: one for video input
    (top) and one for text input (bottom). Given a video and a corresponding text, we encode them to
    frame-level/word-level features. Features belonging to each segment (clip/sentence) are fed to a standard
    temporal transformer (T-Transformer) followed by the proposed feature aggregation module (Attention-FA) to obtain
    clip/sentence-level features. Finally, a new contextual transformer produces the final video/paragraph embedding
    based on interactions between local context (clip/sentence features) and global context (all frames/words
    features). $\ell_{align}^L$, $\ell_{align}^H$, $\ell_{align}^g$ and $\ell_{CMC}$ enforce the model to align the
    representations at different levels.
    }
    \label{fig:fig_model}
\end{figure*}

In this paper, we focus on long-range temporal dependencies and propose a hierarchical model that can exploit
long-range temporal context both in videos and text when learning the joint cross-modal embedding. For instance, the
action of ``making tea'' involves boiling water, pouring it into a cup, and then adding a tea bag. This action can
take a long time and may have lots of details that distinguish a particular style of making tea from other styles. To
capture the whole temporal context, we leverage the idea of a hierarchical model with losses that enforce the
interaction within and between different hierarchy levels. The idea of such a hierarchy is generic and has been
explored by several works~\cite{cmhse,hierarchical_2,hierarchical_3} in the context of video-text learning. In
addition, we use alignment losses from Zhang et al.~\cite{cmhse} and extend our baseline model with a new feature
aggregation method for the intra-level interactions between features and a new transformer-based module for
inter-level interactions (between local and global semantics). We consider three different levels of hierarchy:
frame/word, clip/sentence and video/paragraph, visualized by the three blocks in Figure~\ref{fig:fig_model}.

To model \textit{intra-level cooperation}, we introduce an attention-aware feature aggregation layer to focus on
temporal interactions between low-level entities (Figure~\ref{fig:fig_model}-Attention-FA).

This component replaces traditional sequence representation aggregation methods in transformers such as using a
{\texttt{\small [CLS]} } token~\cite{devlin2018bert,sun2019videobert,tan2019lxmert,su2019vl} or mean
pooling~\cite{kim2019mule} with an attention-aware fusion. It leverages temporal context to encourage important
entities to contribute more to the final representation of a sequence of frames or words.

For the \textit{inter-level cooperation}, we introduce a contextual attention module, which enforces the network to
highlight semantics relevant to the general context of the video and to suppress the irrelevant semantics. This is
done by modeling the interaction between low-level (clips-sentences) and high-level entities (global contexts), as
shown in Figure~\ref{fig:fig_model}-green region.

In addition to this architectural contributions, we introduce a new \textit{cross-modal cycle-consistency loss} to
enforce interaction between modalities and encourage the semantic alignment between them in the learned common space.
We show that enforcing two domains to produce consistent representations leads to substantially improved semantic
alignment.


In summary, this paper contributes:
\begin{itemize}
    \item a hierarchical transformer architecture with a new attention-aware feature aggregation layer and a new
    contextual attention module;
    \item a cross-modal cycle-consistency loss that encourages semantic alignment between vision and text features in
    the joint embedding space;
    \item state-of-the-art results on video-text retrieval.
\end{itemize}

\section{Cooperative Hierarchical Transformer}\label{sec:sec_COOT}
Videos and text descriptions naturally involve different levels of granularity. Every paragraph contains multiple
sentences, and each sentence is composed of several words. Similarly, videos have a hierarchical semantic structure,
even if it is not as exactly defined as for text. Figure~\ref{fig:fig_model}~illustrates the overview of the \coot \
model which consists of three levels:
\emph{1)} A temporal transformer that captures the relationships between frame/word features,
\emph{2)} attention-aware feature aggregation to produce clip/sentence features (Section ~\ref{subsec:intra_c}) and
\emph{3)} a contextual transformer to produce final video and text embeddings (Sec.~\ref{subsec:inter_c}). We use alignment
losses from Zhang et al.~\cite{cmhse} to align representations at different granularity levels. In addition, we
introduce a new cross-model cycle-consistency loss to connect video and text (Sec.~\ref{sec:cross_cc}). In this
section, we briefly summarize the alignment losses from Zhang et al.~\cite{cmhse} and the standard transformer.

\subsection{Preliminaries}\label{subsec:prelim}

\narrowparagraph{Semantic Alignment Losses.}
For the video-text alignment, Zhang et al.~\cite{cmhse} leverage a contrastive loss to enforce the positive samples
to stay in a close neighborhood and negative samples far apart~\cite{contloss06,nofussmetric17,zilos17,ge_metric18}.
Assuming the positive pair $\mathcal{P}=(x, y)$, two negative pairs $(x, y{'})$ and $(x{'}, y)$ expressed
as $\mathcal{N} = \lbrace(x, y{'}),(x{'}, y)\rbrace$, and a margin $\alpha$, they define the following loss:
\begin{equation}
    L(\mathcal{P},\mathcal{N},\alpha) = max(0,\alpha+D(x,y) - D(x{'},y))+max(0,\alpha+D(x,y) - D(x, y{'}))
    \label{eq:lc}
\end{equation}

where $D(x,y) = 1 - x^\intercal y/ (\lVert{x}\rVert\lVert{y}\rVert)$ is the cosine distance of two vectors.

To align representations at clip-sentence ($\vartheta^k_i$, $\delta^k_i$), video-paragraph ($\vartheta^k$,
$\delta^k$) and global context ($g_v$, $g_p$) levels, Zhang et al.~\cite{cmhse} use the following losses:

\begin{equation}
    \begin{split}
        \ell_{align}^L = \underset{{k\in \mathcal D,i,k{'}\neq k,i{'}\neq i}}{\sum}L((\vartheta^k_i,\delta^k_i),
        \lbrace(\vartheta^k_i,  \delta^{k{'}}_{i{'}}),(\vartheta^{k{'}}_{i{'}}, \delta^k_i)\rbrace,\beta)\\
        \ell_{align}^H = \underset{k\in \mathcal D,k{'}\neq k}{\sum}L((\vartheta^k,\delta^k),\lbrace(\vartheta^k,
        \delta^{k{'}}), (\vartheta^{k{'}}, \delta^k)\rbrace,\alpha)\\
        \ell_{align}^g = \underset{k\in \mathcal D,k{'}\neq k}{\sum}L((g_v^k,g_p^k),\lbrace(g_v^k, g_p^{k{'}}),
        (g_v^{k{'}}, g_p^k)\rbrace,\alpha_g)
        \label{eq:l_align}
    \end{split}
\end{equation}

Here, $\vartheta^k_i$ denotes the embedding for the $i$-th clip of the $k$-th video and similarly $\delta^k_i$ is the
embedding of the $i$-th sentence of the $k$-th paragraph. $\alpha$, $\alpha_g$ and $\beta$ are constant margins, and
$\mathcal D$ is a dataset of videos with corresponding text descriptions.
Zhang et al.~\cite{cmhse} employed an additional loss to model the clustering of low-level and high-level semantics
in the joint embedding space:
\begin{equation}
    \begin{split}
        \ell_{cluster} = \underset{{k\in \mathcal D,i,k{'}\neq k,i{'}\neq i}}{\sum}L((1,1)&,\lbrace(\vartheta^k_i,
        \vartheta^{k{'}}_{i{'}}), (\delta^{k{'}}_{i{'}}, \delta^k_i)\rbrace,\gamma)\\
        &+
        \underset{k\in \mathcal D,k{'}\neq k}{\sum}L((1,1),\lbrace(\vartheta^k, \vartheta^{k{'}}), (\delta^{k{'}},
        \delta^k)\rbrace,\eta)
        \label{eq:l_cluster}
    \end{split}
\end{equation}
\noindent where $\gamma$ and $\eta$ both are constant margins. The $(1, 1)$ pairs denote that positive samples are
not changed. In short, the goal of this loss is to push apart embeddings for negative samples.

\narrowparagraph{Note.} Due to the symmetrical design of the video and text branches in our model, from now on, we explain
only the video branch. For simplicity, we assume a single head in transformer formulations. All transformers use
residual connections.

\narrowparagraph{Temporal Transformer.}
We use standard attention-blocks~\cite{attention_nips17} to learn frame and word representations, as shown in
Fig~\ref{fig:fig_model}-Right.
We learn two temporal transformers (T-Transformer); one for the video branch and another one for the text branch.
Both have the same architecture. All T-Transformers in each branch share their weights. This module draws the
relationship between temporal features and yields improved representations as output.
Given a video $v^k$, we first encode all its frames to obtain the frame-level features $\{f^k_{i,:} \}_{i=1}^n$,
where $f^k_{i,:}$ are all frame-level features of the $i$-th clip for video $v^k$ (orange parts in
Figure~\ref{fig:fig_model}). We also consider all frame features ($f^k_{:}$) of a video as extra input for the global
context computation (green parts in Figure~\ref{fig:fig_model}). This yields $\{\hat{f}^k_{i,:} \}_{i=1}^n$ and
$\hat{f}^k_{:}$.


\subsection{Intra-Level Cooperation}\label{subsec:intra_c}
Standard feature fusion methods consider each feature independently by average pooling or max pooling. Hence, they
miss the relationship between features to highlight the relevant features. Recent transformers use a {\texttt{\small
[CLS]}} token~\cite{sun2019videobert,devlin2018bert,tan2019lxmert,su2019vl} or average pooling~\cite{kim2019mule} to
obtain the aggregated features.
For example, when a person is cooking, objects on the table are more relevant than objects on the wall or in the
background. Therefore, we need to attend to specific features depending on the context. There have been some attempts
in other domains to design a context-aware feature fusion method~\cite{pooling2,pooling3,pooling1,pooling4}. However,
we introduce an attention-aware feature aggregation module (Attention-FA in Fig.~\ref{fig:fig_model}) for video-text
transformers.

Suppose we have a sequence with $T$ feature vectors, denoted by $X=\{x_1,\dots,x_T\}$ (e.g. $\hat{f}^k_{i,:}  =
\{\hat{f}^k_{i,1},\dots,\hat{f}^k_{i,T}\}$). We set \textbf{key} $K=X$ and utilize two learnable transformation
weights $W_2$ and $W_1$ together with two biases $b_1$ and $b_2$. The attention matrix $A$ is computed as:
\begin{equation}
    A = \text{softmax}(W_2 Q+b_2)^T, \quad \quad Q=GELU(W_1 K^T+b_1),\ K=X
    \label{eq:equation_softmax}
\end{equation}
We compute the final feature as $\hat{x} = \sum_{i=1}^{T}a_i\odot x_i$, where $\odot$ denotes element-wise
multiplication and $a_i$ is the $i$-th attention vector of $A$ for the $i$-th feature. This module differs from
attention~\cite{attention_nips17} in two aspects: (1) we use only two learnable weights for query ($Q$) and key ($K$)
and then aggregate the values based on calculated scores; (2) the query equals to transformed keys (K) and then we
apply the activation function GELU~\cite{gelu1,gelu2}. We feed $\{\hat{f}^k_{i,:} \}_{i=1}^n$ and  $\hat{f}^k_{:}$ to
this component and obtain the clip-level ($\{\vartheta^k_{i} \}_{i=1}^n$) features and the global context for the
video ($g_\nu$).

\begin{figure*}[t]
    \centering
    \includegraphics[width=1.0\textwidth]{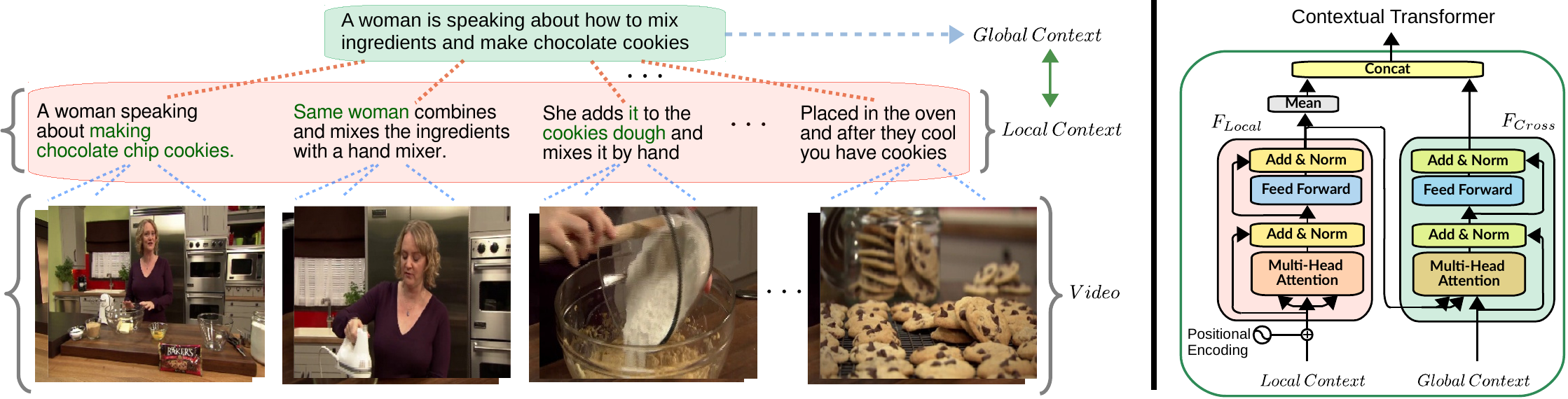}
    \caption{
    \textbf{Contextual Transformer (CoT).}
    This module (right) encourages the model to optimize the representations with respect to interactions between
    local and global context. In the third sentence, to know the type of dough (\textit{cookie}) the model should
    have information about the general context of the video (\textit{making chocolate cookies}). Likewise, in the
    second sentence, to know that she is the \textit{"same woman"}, the model must be aware of the person's identity
    throughout the video.
    }
    \label{fig:lgt}
\end{figure*}

\subsection{Inter-Level Cooperation}\label{subsec:inter_c}

By modeling the interactions between local and global context, the network learns to highlight semantics relevant to
the general context of the video and to suppress the irrelevant ones: interactions between clip embeddings and the
general context of the video; interactions between sentence embeddings and the general context of the text.
As shown in Figure~\ref{fig:lgt}-Left, without knowing the global context, just from observing the frame in the third
clip, there is no information about what type of \textit{"dough"} is involved. Also the \textit{"same woman"} in the
second clip could not be related to the woman seen in the first clip.

Thus, we propose a Contextual Transformer (CoT) in Figure~\ref{fig:lgt}-Right to model the interactions between
low-level and high-level semantics.
More formally, we build the Contextual Transformer with two modules $F_{Local}$ and $F_{Global}$. We append the
positional embedding to the inputs of $F_{Local}$. The goal of $F_{Local}$ is to model the short-term interactions
between low-level semantics ($\{\vartheta^k_{i} \}_{i=1}^n$), whereas $F_{Global}$ models the interactions between
local and global context ($g_\nu$) to highlight the important semantics.

Given local representations $\{\vartheta^k_i\}_{i=1}^n \in \mathbb{R}^{n\times d}$, where $n$ is the number of clips
and $d$ indicates the feature dimension, $F_{Local}$ applies multi-head attention followed by a feed-forward layer
and a normalization layer on top of both layers and produces embeddings $\{h_i\}_{i=1}^n$.

We compute \textbf{key (K)-value(V)} pairs based on these embeddings $\{h_i\}_{i=1}^n \in \mathbb{R}^{n\times d}$ and
\textbf{query(Q)} based on the global context $g_v$. $F_{Global}$  produces the attention output as follows,
\begin{equation}
    H_{attn} = \text{softmax}(\frac{\textbf{QK}^T}{\sqrt{d}})\textbf{V},\quad \quad Q=\mathcal{W}_q g_v,\
    K=\mathcal{W}_k\{h_i\}_{i=1}^n,\ V=\mathcal{W}_v\{h_i\}_{i=1}^n
    \label{eq:equation_attention}
\end{equation}

where $\mathcal{W}_q$, $\mathcal{W}_k$, and $\mathcal{W}_v$ are the embedding weights. $H_{attn}$ is a weighted sum
of values (local semantics), where the weight of each value is calculated based on its interaction with the global
context query \textbf{Q}. $H_{attn}$ is further encoded by a feed-forward layer to produce the contextual embedding
$H_{context}$. We calculate the mean of $\{h_i\}_{i=1}^n$ and concatenate it with $H_{context}$ to obtain the final
video embedding $\vartheta^k = \text{concat}(mean(\{h_i\}_{i=1}^n),H_{context})$; see Figure~\ref{fig:lgt}.


\section{Cross-Modal Cycle Consistency}
\label{sec:cross_cc}
\begin{wrapfigure}{r}{0.5\textwidth}
    \centering
    \includegraphics[width=0.5\textwidth]{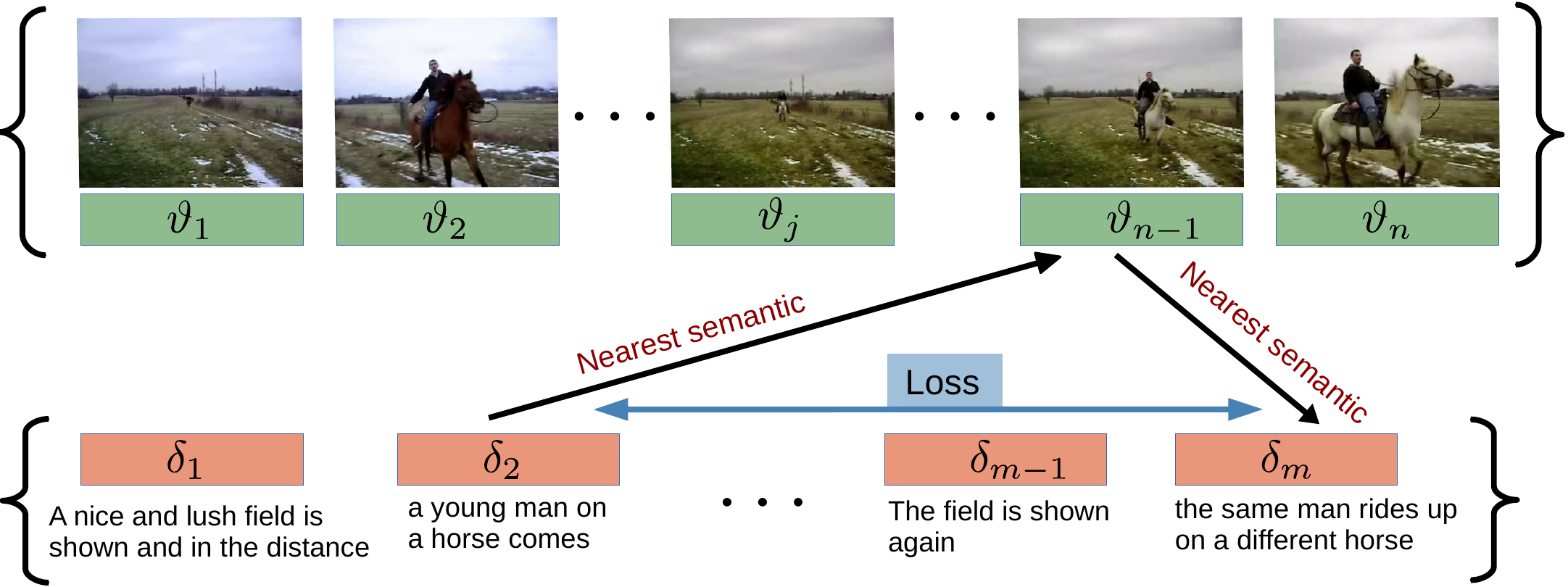}
    \caption{\textbf{Cross-Modality Cycle-Consistency.} Starting from a sentence $s_i$, we find its nearest neighbor
    in the clip sequence and again its neighbor in the sentence sequence. Deviations from the start index are
    penalized as alignment error.}
    \vspace{-3mm}
    \label{fig:ccloss}
\end{wrapfigure}

We introduce a cross-modal cycle-consistency loss to enforce the semantic alignment between clips and sentences, as
illustrated in Figure~\ref{fig:ccloss}. It replaces the cross-modal attention units used in~\cite{tan2019lxmert,
actbert20}. A pair of clip and sentence will be identified as semantically aligned if they are nearest neighbors in
the learned common spaces. Consider as input a sequence of clip embeddings $\{\vartheta_i\}_{i=1}^n =\{\vartheta_1,
\dots,\vartheta_n\}$ and sentence embeddings $\{\delta_i\}_{i=1}^m=\{\delta_1,\dots,\delta_m\}$.
As the sentences of a paragraph have a temporal order, given a sentence embedding $\delta_i$ on this sequence, we
first find its soft nearest neighbor~\cite{Rocco18b,NIPS2004_2566,cycle_consistency1}
\begin{equation}
    \bar{\vartheta}_{\delta_i} = \sum_{j=1}^{n}\alpha_j\vartheta_j \,\,\,\,\,\, \text{where}\,\,\,\,\,\,
    \alpha_j=\frac{\exp(-\lVert \delta_i-\vartheta_j\rVert^2)}{\sum_{k=1}^{n}\exp(-\lVert \delta_i-\vartheta_k\rVert^2)}
    \label{eq:equation_cc1}
\end{equation}
in the clip sequence $\{\vartheta_i\}_{i=1}^n$. $\alpha_j$ is the similarity score of clip $\vartheta_j$ to sentence
$\delta_i$.
We then cycle back from $\bar{\vartheta}_{\delta_i}$ to the sentence sequence $\{\delta_i\}_{i=1}^m$ and calculate
the soft location
\begin{equation}
    \mu = \sum_{j=1}^{m}\beta_j j \,\,\,\,\,\, \text{where}\,\,\,\,\,\, \beta_j=\frac{\exp(-\lVert
    \bar{\vartheta}-\delta_j\rVert^2)}{\sum_{k=1}^{m}\exp(-\lVert \bar{\vartheta}-\delta_k\rVert^2)}.
    \label{eq:equation_cc2}
\end{equation}
The sentence embedding $\delta_i$ is semantically cycle consistent if and only if it cycles back to the original
location, i.e., $i=\mu$. We penalize deviations from cycle-consistency for sampled sets of clips and sentences, which
encourages the model to learn semantically consistent representations.

Our objective is the distance between the source location $i$ and the soft destination location $\mu$.
\begin{equation}
    \ell_{CMC} = \lVert i-\mu\rVert^2
    \label{eq:equation_cc3}
\end{equation}
Computing nearest neighbors as soft nearest neighbors makes the loss differentiable~\cite{Rocco18b,NIPS2004_2566,
cycle_consistency1}.
We can use this loss in both supervised and self-supervised scenarios. In the self-supervised case, we split each
video uniformly into several clips and each paragraph into sentences. Beside the cycle-consistency from text to
video, we also calculate $\ell_{CMC}$ from video to text. Therefore, the final $\ell_{CMC}$ loss includes both cycles.

The final training loss for the overall model is:
\begin{equation}
    \ell_{final} = \ell_{align}^L+\ell_{align}^H+\ell_{align}^g + \ell_{cluster} + \lambda\ell_{CMC}
    \label{eq:equation_total_loss}
\end{equation}



    \section{Experimental Setup}\label{sec:experimental-setup}

\narrowparagraph{Datasets.}
We evaluate our method on the datasets \textbf{ActivityNet-captions}~\cite{krishna2017dense}
and \textbf{Youcook2}~\cite{ZhXuCoCVPR18}. ActivityNet-captions consists of 20k YouTube videos with an average length
of 2 minutes, with 72k clip-sentence pairs. There are $\sim$10k, $\sim$5k and $\sim$5k videos in train, val1 and
val2, respectively.
Youcook2 contains 2000 videos with a total number of 14k clips. This dataset is collected from YouTube and covers 89
types of recipes. There are $\sim$9.6k clips for training and $\sim$3.2k clips for validation. For each clip there is
a manually annotated textual description.

\narrowparagraph{Evaluation Metrics.} We measure the performance on the retrieval task with standard retrieval metrics, i.e.,
recall at K (R$@$K e.g.\ R$@$1, R$@$5, R$@$10) and Median Rank (MR).

\label{sec:imp_details}

\narrowparagraph{Text encoding.}
We feed paragraphs consisting of several sentences into a pretrained "BERT-Base, Uncased" model~\cite{devlin2018bert}
and use the per-token outputs of the last 2 layers, resulting in 1536-d features.

\narrowparagraph{Video encoding.}
For Activitynet-Captions, we use the 2048-d features provided by Zhang et al.~\cite{cmhse} (at 3.8 FPS). For
Youcook2, we test two approaches: (\textit{A}) We follow Miech et al., 2019 ~\cite{miech19howto100m} and concatenate
2D (Resnet-152 pretrained on ImageNet~\cite{resnetimagenet}) and 3D (ResNext-101 model~\cite{Hara2018CanS3}
pretrained on Kinetics~\cite{resnextkinetics}) outputs to obtain 4096-d features at 3 FPS; (\textit{B}) We use the
video embedding network provided by Miech et al., 2020~\cite{miech20endtoend} pretrained on video-text learning on
the Howto100m dataset to obtain 512-d features at 0.6 FPS\@.

For each clip as well as for the entire video, we sample up to 80 frame features. If needed, we split the frames into
80 equal length intervals and uniformly sample a frame from each interval during training or take the center frame
during validation.

\narrowparagraph{Training.}
Similar to~\cite{cmhse} we set all margins $\alpha=\alpha_g=\beta=\gamma=\mu=0.2$. We use a mini-batch size of 64
video/paragraph pairs and sample all corresponding clips and sentences. All possible combinations of embeddings with
non-matching indices in a batch are used as negative samples for the contrastive loss. To apply the cycle-consistency
loss, we found that sampling 1 clip per video and 1 sentence per paragraph works best. The optimal loss weight
$\lambda$ depends on architecture and dataset.
As activation function, we found GELU~\cite{gelu1} to perform best. We set the hidden size to 384 and use a pointwise
linear layer to reduce the input feature dimension.
We use one self-attention layer for the T-Transformer and one self-attention and one cross-attention layer for CoT\@.
For further details on optimization and hyperparameters we refer the interested reader to the supplementary material.

\narrowparagraph{Video-Language Retrieval.}
For video-text retrieval, the query is a paragraph and the task is to find the most relevant video from a database.
Alternatively, the query can be a video and the task is to retrieve the most relevant paragraph. We follow the
experimental protocol from Zhang et al.~\cite{cmhse} to evaluate the models. We use the final embedding output of our
model ($\vartheta^k, \delta^k$) to do the retrieval.

\narrowparagraph{Clip-sentence retrieval.}
For Youcook2, we also evaluate the quality of our model when retrieving a short video clip given a single sentence.
For this experiment, we use the intermediate low-level embeddings produced by our model ($\vartheta^k_i, \delta^k_i$)
to do the retrieval.

    \section{Results}\label{sec:results}


\narrowparagraph{Influence of each component.}
We show results of a model ablation study in Table~\ref{tab:ab_components}.
First, to validate the general effectiveness of the proposed cross-modal cycle consistency loss (CMC), we apply it to
the HSE architecture~\cite{cmhse}. The $\ell_{CMC}$ loss provides a significant boost in performance for both HSE and
\coot, which indicates that it will be beneficial if plugged into other video-text representation learning methods.
Second, the Attention-FA module shows better performance (7.2\% average improvement on R$@$1 for
paragraph$\implies$video and video$\implies$paragraph tasks) than common average pooling.
Third, we observe that integrating the Contextual Transformer into the overall model improves the performance. This
confirms that interactions between local and global context help the model to highlight the relevant semantics (more
in supp. material).
\begin{table}[t]
    \caption{\textbf{Ablation study on ActivityNet-captions (val1).} We quantify the individual contributions of the
    attention-aware feature aggregation (AF), the Contextual Transformer (CoT), and the cross-modal
    cycle-consistency loss (CMC). HSE results are reproduced by us. Disabling CoT means removing the
    cross-attention layer between local and global context.}
    \centering
    \small
    \setlength{\tabcolsep}{0.5em}
    \resizebox{1\columnwidth}{!}{
    \begin{tabular*}{\textwidth}{llcccccccccc}
        \thickhline
        \multirow{2}{*}{Model} & Pooling & \multirow{2}{*}{CMC} & \multirow{2}{*}{CoT} &
        \multicolumn{3}{c}{Paragraph$\implies$Video} & \multicolumn{3}{c}{Video$\implies$Paragraph} &
        \multirow{2}{*}{Param (M)}\\
        \cmidrule(rl){2-2} \cmidrule(rl){5-7}  \cmidrule(rl){8-10}
        & Lowlvl & & & R@1 & R@5 & R@50  & R@1 & R@5 & R@50  & \\ \thickhline
        
        
        HSE   & Max & \xmark & \xmark & 45.6\std{0.3} & 76.1\std{0.7} & 96.0\std{0.3} & 44.9\std{0.5} & 75.8\std{1.2} & 95.8\std{0.4} & 26.1 \\
        HSE   & Max & \cmark & \xmark & 46.6\std{0.4} & 78.1\std{0.3} & 97.3\std{0.1} & 46.4\std{0.3} & 77.6\std{0.3} & 97.1\std{0.3} & 26.1 \\ \hline
        
        \coot & CLS & \xmark & \xmark & 49.4\std{1.4} & 77.7\std{1.3} & 95.7\std{0.2} & 49.7\std{1.9} & 77.8\std{0.9} & 95.8\std{0.3} & 4.9  \\
        \coot & AVG & \xmark & \xmark & 52.6\std{0.6} & 80.6\std{0.4} & 97.0\std{0.2} & 52.1\std{0.4} & 80.8\std{0.2} & 97.0\std{0.2} & 4.9  \\
        \coot & Max & \xmark & \xmark & 58.2\std{0.5} & 84.9\std{0.2} & 98.1\std{0.1} & 58.7\std{0.5} & 86.0\std{0.2} & 98.2\std{0.1} & 4.9  \\
        \coot & AFA & \xmark & \xmark & 59.0\std{0.5} & 85.4\std{0.2} & 98.2\std{0.0} & 59.8\std{0.6} & 85.8\std{0.8} & 98.2\std{0.1} & 5.8  \\
        \hline
        \coot & Max & \cmark & \cmark & 59.4\std{0.9} & 86.1\std{0.6} & 98.3\std{0.0} & 60.5\std{0.1} & 87.1\std{0.2} & 98.5\std{0.1} & 6.7  \\
        \coot & AFA & \xmark & \cmark & 59.8\std{1.1} & 86.3\std{0.3} & 98.5\std{0.1} & 60.1\std{0.1} & 87.1\std{0.4} & 98.5\std{0.1} & 7.6  \\
        \coot & AFA & \cmark & \xmark & 59.5\std{0.5} & 85.5\std{0.4} & 98.1\std{0.0} & 60.5\std{0.7} & 86.2\std{0.5} & 98.2\std{0.1} & 5.8  \\ \hline
        
        \coot & AFA & \cmark & \cmark & \textbf{60.8}\std{0.6} & \textbf{86.6}\std{0.4} & \textbf{98.6}\std{0.1} & \textbf{60.9}\std{0.3} & \textbf{87.4}\std{0.5} & \textbf{98.6}\std{0.0} & 7.6 \\
        
        
        \thickhline
    \end{tabular*}
    }
    
    \label{tab:ab_components}
\end{table}

\narrowparagraph{Comparison to the state of the art.}
Table~\ref{tab:activitynet} summarizes the results of paragraph to video and video to paragraph retrieval tasks on
the ActivityNet-captions dataset. For a fair comparison, our model utilizes the same video features as HSE~\cite{cmhse}.
Our method significantly outperforms all previous methods across different evaluation metrics.
\coot \ obtains on average ~16.6\% better R$@$1 in comparison to HSE~\cite{cmhse} while having fewer parameters. We
believe the major gain comes from our attention-aware feature aggregation component and the $\ell_{CMC}$ loss.

We further provide retrieval results on the Youcook2~\cite{ZhXuCoCVPR18} dataset in Table~\ref{tab:youcook2}.
We compare our model under two settings: (1) with features pretrained on classification (2) with features from a
pretrained SOTA video-text model.

\emph{Without HowTo100M pretrained features.} We use features (\textit{A}) explained in
Section~\ref{sec:imp_details} and train the \coot \ model on the YouCook2 dataset. Using the same training set, \coot
\ outperforms Miech et al.~\cite{miech19howto100m} and HGLMM~\cite{klein15} on both paragraph-to-video and
sentence-to-clip tasks. This supports our rationale that modeling interactions between different hierarchy levels is
crucial for capturing long-term semantics.

\emph{With HowTo100M pretrained features.}
In Table~\ref{tab:youcook2}, we compare our method with the recently proposed SOTA methods
MIL-NCE~\cite{miech20endtoend}, ActBERT~\cite{actbert20}, and Miech et al.~\cite{miech19howto100m}, which utilize
pretraining on the huge HowTo100M dataset. We use features (\textit{B}) (Sec.~\ref{sec:imp_details}) and train the
model on the YouCook2 dataset.
Note that the paragraph to video results of other methods are computed by us.
Training our model with features of a model pretrained on the HowTo100M dataset clearly improves over training with
features of a model pretrained on classification and over the state-of-the-art. We can see that our model outperforms
MIL-NCE~\cite{miech20endtoend} 16.4\% on R$@$1 score for paragraph-to-video task, which verifies that \coot \ benefits
from hierarchy interactions.
This shows that the contributions of this paper are complementary to works that focus on large-scale pretraining.

\narrowparagraph{Time complexity and number of parameters.} The \coot \ model has 10.6M, parameters which is 60\% less than
the HSE method (Table~\ref{tab:ab_components}). Training is fast and takes less than 3 hours on two GTX1080Ti GPUs
(without data I/O).


\subsection{Video Captioning}\label{subsec:cap}

To show that the learned representations contain meaningful information for other tasks than retrieval,
we use the learned representations for video captioning building upon the captioning model MART~\cite{mart}.
The original method uses appearance (RGB) and optical flow features extracted from ResNet-200~\cite{resnetimagenet}
and BN-Inception~\cite{bninception}, respectively.

We use the clip ($\vartheta^k_i$) and optionally the video ($\vartheta^k$) representation generated with our COOT model.
In comparison to MART, we input about 100 times less features per video into the captioning model.
We use the standard language evaluation metrics BLEU@3/4~\cite{papineni-etal-2002-bleu},
RougeL~\cite{lin-2004-rouge}, METEOR~\cite{denkowski-lavie-2014-meteor}, CIDEr-D~\cite{cider} and
R$@$4~\cite{ngramrepetition} which measures the degree of n-gram repetition.
Our results in Table~\ref{tab:captioning_yc} and Table~\ref{tab:captioning_an} show that the MART method using our
representations improves over
using appearance and optical flow video features.
Generated captions in Table~\ref{tab:cap_anettest} show that our video representations encapsulate richer information
about the video while being more compact.
\begin{table}[t]
    \caption{\textbf{Video-paragraph retrieval results on AcitvityNet-captions dataset (val1).}}\vspace{5pt}
    \centering
    \footnotesize
    \resizebox{1\columnwidth}{!}{
    \begin{tabular}{l c c c c  | c c c c }
        \thickhline
        \multicolumn{1}{c}{} &
        \multicolumn{4}{c}{Paragraph$\implies$Video} &
        \multicolumn{4}{c}{Video$\implies$Paragraph}   \\
        \cmidrule(r){2-5}
        \cmidrule(r){6-9}
        Method                                              & R$@$1             & R$@$5             & R$@$50            & MR            & R$@$1             & R$@$5             & R$@$50            & MR            \\ \thickhline
        LSTM-YT~\cite{venugopalan2015sequence}              & 0.0               & 4.0               & 24.0              & 102.0         & 0.0               & 7.0               & 38.0              & 98.0          \\
        No Context~\cite{venugopalan2014translating}        & 5.0               & 14.0              & 32.0              & 78.0          & 7.0               & 18.0              & 45.0              & 56.0          \\
        DENSE~\cite{krishna2017dense}                       & 14.0              & 32.0              & 65.0              & 34.0          & 18.0              & 36.0              & 74.0              & 32.0          \\
        VSE~\cite{NIPS2013_5204}(~\cite{Shao_2018_ECCV})    & 11.7              & 34.7              & 85.7              & 10            & -                 & -                 & -                 & -             \\
        FSE~\cite{cmhse}                                    & 18.2              & 44.8              & 89.1              & 7             & 16.7              & 43.1              &  88.4             & $7$           \\
        HSE~\cite{cmhse}                                    & \nsn{44.4}{0.5}   & \nsn{76.7}{0.3}   & \nsn{97.1}{0.1}   & 2             & \nsn{44.2}{0.6}   & \nsn{76.7}{0.3}   & \nsn{97.0}{0.3}   & 2             \\ \hline
        COOT                                                & \nsb{60.8}{0.6}   & \nsb{86.6}{0.4}   & \nsb{98.6}{0.1}   & \textbf{1}    & \nsb{60.9}{0.3}   & \nsb{87.4}{0.5}   & \nsb{98.6}{0.0}   & \textbf{1}    \\ \thickhline
    \end{tabular}}
    \label{tab:activitynet}
\end{table}

\begin{table}[t]
    \caption{
        \tbf{Retrieval results on YouCook2 dataset.}
        Results with * are computed by us. $^\bigtriangleup$ we use features of a video-text
        model~\cite{miech20endtoend} pretrained on the HowTo100m dataset.}
    \centering
    \setlength\tabcolsep{.3em}
    \resizebox{1\columnwidth}{!}{
    \renewcommand*{\arraystretch}{1.1}
    \begin{tabular}{lll llll llll}
        \thickhline
        & & \multicolumn{4}{c}{Paragraph$\implies$Video} & \multicolumn{4}{c}{Sentence$\implies$Clip} \\
        \cmidrule(r){3-6} \cmidrule(r){7-10}
         Method                                    & TrainSet         & R$@$1             & R$@$5             & R$@$10            & MR                & R$@$1             & R$@$5             & R$@$10            & MR                \\ \thickhline
         Random                                    & -                & 0.21              & 1.09              & 2.19              & 229               & 0.03              & 0.15              & 0.3               & 1675              \\
         Miech et al.~\cit{miech19howto100m}       & HowTo100M        & 43.1*             & 68.6*             & 79.1*             & 2*                & 6.1               & 17.3              & 24.8              & 46                \\
         ActBERT~\cit{actbert20}                   & HowTo100M        & -                 & -                 & -                 & -                 & 9.6               & 26.7              & 38.0              & 19                \\
        MIL-NCE~\cit{miech20endtoend}             & HowTo100M        & 61.9*             & 89.4*             & 98.9*             & \tbf{1}*          & 15.1              & 38.0              & 51.2              & 10                \\ \hline
         HGLMM~\cit{klein15}                       & YouCook2         & -                 & -                 & -                 & -                 & 4.6               & 14.3              & 21.6              & 75                \\
    Miech et al.~\cit{miech19howto100m}       & YouCook2         & 32.3*             & 59.2*             & 70.9*             & 4*                & 4.2               & 13.7              & 21.5              & 65                \\
         COOT                                      & YouCook2         & \nsn{50.4}{2.6}   & \nsn{79.4}{0.6}   & \nsn{87.4}{0.8}   & \nsn{1.3}{0.6}    & \nsn{5.9}{0.7}    & \nsn{16.7}{0.6}   & \nsn{24.8}{0.8}   & \nsn{49.7}{2.9}   \\ \hline
        \mr{Miech et al.~\cit{miech19howto100m}}  & HowTo100M+       & \mr{59.6*}        & \mr{86.0*}        & \mr{93.6*}        & \mr{\tbf{1}*}     & \mr{8.2}          & \mr{24.5}         & \mr{35.3}         & \mr{24}           \\
                        &                                            YouCook2         &                   &                   &                   &                   &                   &                   &                   &                   \\
          \mr{COOT}                                 & HowTo100M$^\bigtriangleup$+   & \msb{77.2}{1.0}   & \msb{95.8}{0.8}   & \msb{97.5}{0.3}   & \msb{1.0}{0.0}    & \msb{16.7}{0.4}   & \msb{40.2}{0.3}   & \msb{52.3}{0.5}   & \msb{9.0}{0.0}    \\
        &                                                            YouCook2         &                   &                   &                   &                   &                   &                   &                   &                   \\ \thickhline
    \end{tabular}
    } 
    \label{tab:youcook2}
\end{table}

%

\newcommand{\howm}{H100M$^\bigtriangleup$+YC2} 
\newcommand{\cootclip}{\coot~clip}
\newcommand{\cootvidclip}{\coot~video+clip}
\newcommand{\cootvid}{\coot~video only}
\newcommand{\martfeats}{RGB+Flow}

\begin{table}[t]
    \caption{
        \tbf{Captioning results on the YouCook2 dataset (val split).}
        Results with * are computed by us. $^\bigtriangleup$ we use features of a video-text
        model~\cite{miech20endtoend} pretrained on the HowTo100m dataset.
        "MART w/o re" denotes a MART variant without recurrence.
        }
    \centering

    \resizebox{1\columnwidth}{!}{
    \renewcommand*{\arraystretch}{1.1}
    \begin{tabular}{lll llll llll}
        \thickhline
Features        & Method            & TrainSet  & B$@$3     & B$@$4     & RougeL   & METEOR         & CIDEr-D      & R$@$4$\downarrow$\\\thickhline
\martfeats   & VTransformer~\cite{zhou2018endtoend}      & YouCook2  &     13.08*&      7.62 &     32.18*&     15.65 &     32.26 &      7.83 \\
\martfeats   & TransformerXL~\cite{dai2019transformerxl}     & YouCook2  &     11.46*&      6.56 &     30.78*&     14.76 &     26.35 &      6.30 \\
\martfeats   & MART~\cite{mart}              & YouCook2  &     12.83*&      8.00 &     31.97*&     15.90 &     35.74 &\tbf{4.39}\\
\hline
\cootclip       & MART              & YouCook2  &     14.17 &      8.69 &     33.01 &     16.11 &     38.28 &      8.07 \\
\cootvidclip    & MART              & YouCook2  &     15.75 &      9.44 &     34.32 &     18.17 &     46.06 &      6.30 \\
\hline
\cootclip       & MART              & \howm     &     17.12 &     10.91 &     37.59 &     18.85 &     54.07 &      5.11 \\
\cootclip       & MART w/o re.      & \howm     &     17.16 &     10.69 &     37.43 &     19.18 &     54.85 &      5.45 \\
\cootclip       & VTransformer      & \howm      &    17.62 &     11.09 &     37.63 &     19.34 &     54.67 &      4.57 \\
\hline
\cootvidclip    & VTransformer~\cite{zhou2018endtoend}      & \howm     &     17.79 &     11.05 &     37.51 &     19.79 &     55.57 &      5.69 \\
\cootvidclip    & MART              & \howm     &\tbf{17.97}&\tbf{11.30}&\tbf{37.94}&\tbf{19.85}&\tbf{57.24}&      6.69 \\
\hline
\thickhline
    \end{tabular}
    } 
    \label{tab:captioning_yc}
\end{table}

\begin{table}[t]
    \caption{
        \tbf{Captioning results on the ActivityNet-Captions dataset (ae-test split of MART~\cite{mart}).}
        Results with * are computed by us.
        "MART w/o re" denotes a MART variant without recurrence.}
    \centering

    \resizebox{1\columnwidth}{!}{
    \renewcommand*{\arraystretch}{1.1}
    \begin{tabular}{lll llll llll}
        \thickhline
Features        & Method            & TrainSet    & B$@$3     & B$@$4     & RougeL   & METEOR         & CIDEr-D         & R$@$4$\downarrow$\\\thickhline
\martfeats   & VTransformer~\cite{zhou2018endtoend}      & ActivityNet &     16.27*&      9.31 &     29.18*&     15.54 &     21.33 &      7.45 \\
\martfeats   & TransformerXL~\cite{dai2019transformerxl}     & ActivityNet &     16.71*&     10.25 &     30.53*&     14.91 &     21.71 &      8.79 \\
\martfeats   & MART              & ActivityNet &     16.43*&     9.78  &     30.63*&     15.57 &     22.16 &      5.44 \\
\hline

\cootvidclip    & TransformerXL~\cite{dai2019transformerxl}     & ActivityNet &     16.94 &     10.57 &     30.93 &     14.76 &     22.04 &     15.85 \\
\cootvidclip    & VTransformer~\cite{zhou2018endtoend}      & ActivityNet &     16.80 &     10.47 &     30.37 &     15.76 &     25.90 &     19.14 \\
\cootclip       & MART w/o re.      & ActivityNet &     15.41 &      9.37 &     28.66 &     15.61 &     22.05 &     12.03 \\
\cootvidclip    & MART w/o re.      & ActivityNet &     16.59 &     10.33 &     29.93 &     15.64 &     25.41 &     17.03 \\\hline
\cootclip       & MART              & ActivityNet &     16.53 &     10.22 &     30.68 &     15.91 &     23.98 & \tbf{5.35}\\
\cootvidclip    & MART              & ActivityNet &\tbf{17.43}&\tbf{10.85}&\tbf{31.45}&\tbf{15.99}&\tbf{28.19}&      6.64 \\

\thickhline
    \end{tabular}
    } 
    \label{tab:captioning_an}
\end{table}

\definecolor{myrept}{rgb}{0.1, 0.1, 0.8}
\newcommand{\rept}[1]{\textit{\textcolor{myrept}{#1}}}
\definecolor{mycont}{rgb}{0.8, 0.1, 0.1}
\newcommand{\cont}[1]{\textbf{\textcolor{mycont}{#1}}}

\begin{table}[t]
    \caption{
    \textbf{Captioning samples, more accurate (left) and less accurate (right) cases.}
    First row: ActivityNet (ae-test split), second row: YouCook2 (val split).
    Red/bold indicates content errors, blue/italic indicates repetitive patterns.
    }
    \centering
    \small
    \setlength\tabcolsep{0.3em} 
    \renewcommand*{\arraystretch}{1.1} 
    \begin{tabularx}{\linewidth}{XX}
        \includegraphics[width=\linewidth]{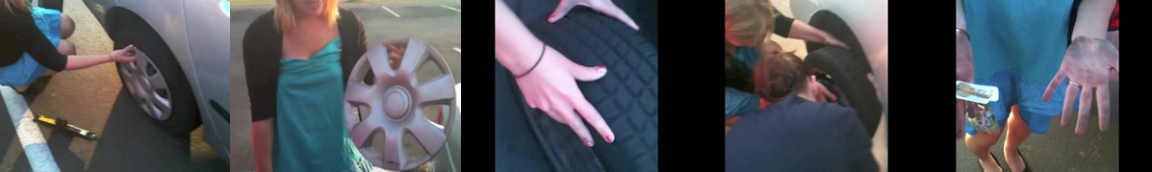} &
        \includegraphics[width=\linewidth]{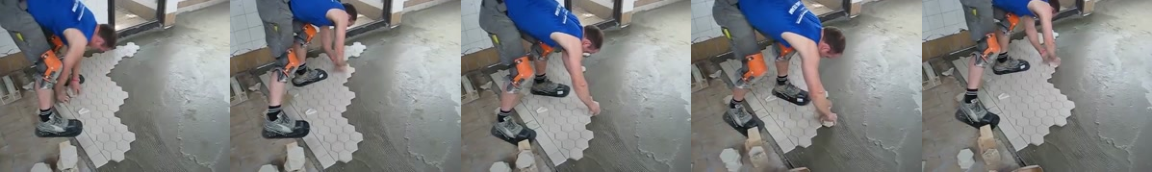} \\
        \begin{minipage}[t]{0.48\textwidth}
            \textbf{MART}: A person is \cont{driving the car}. \cont{A boy} is \cont{holding a bottle of wood}. \\
            \textbf{COOT (Ours)}: A woman is seen kneeling down next to a car while others stand around her.
            The woman then pushes the tire \cont{back and fourth}. \\
            \textbf{GT}: A girl is shown trying to change a tire. She successfully removes the tire,
            then replaces it with a spare, showing off their dirty hands afterward. \\
        \end{minipage}
        &
        \begin{minipage}[t]{0.48\textwidth}
            \textbf{MART}: A man \rept{is kneeling down on} a floor. He \rept{is kneeling down on} the ground. \\
            \textbf{COOT (Ours)}: A man is seen kneeling down on the ground and begins \cont{putting shoes on}.
            The man continues to \rept{put on the shoes} and ends by \rept{putting his shoes on}. \\
            \textbf{GT}: A person is seen bending over a floor placing tiles down over the plaster.
            The person continues laying tiles down and pushing down on the floor to make sure it's sturdy. \\
        \end{minipage}
        \\
        \includegraphics[width=\linewidth]{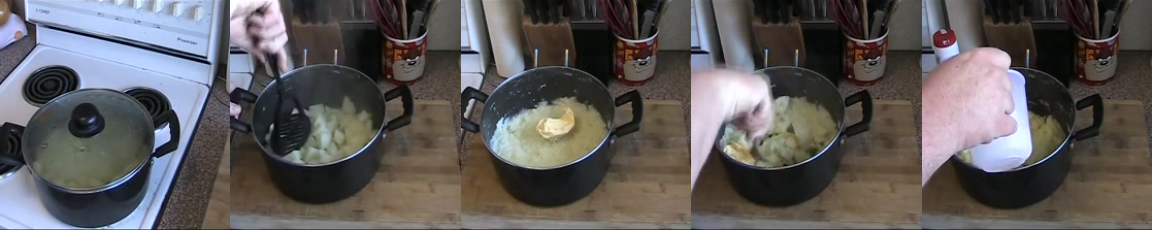} &
        \includegraphics[width=\linewidth]{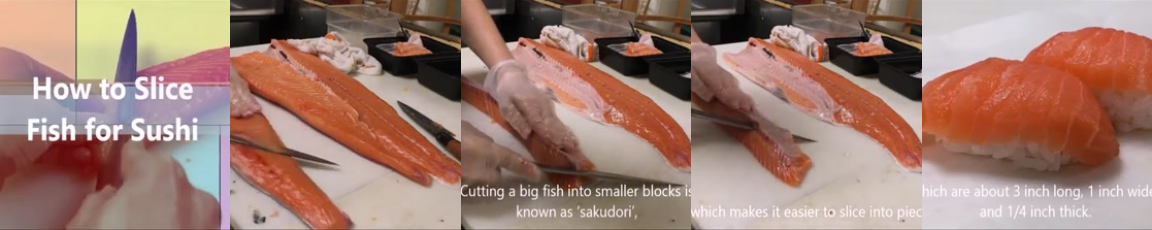} \\
        \begin{minipage}[t]{0.48\textwidth}
            \textbf{MART}: Heat up a pan and \cont{cook until golden brown}. Add \cont{onions} \rept{to the pan}.
            Add \cont{flour} salt and \cont{pepper} \rept{to the pan}. Add \cont{rice} \rept{to the pan} and stir. \\
            \textbf{COOT (Ours)}: Boil the potatoes in water. Add chopped \rept{potatoes} to the pan.
            Add butter and mash. Add some milk and \rept{mash}. \\
            \textbf{GT}: Boil some small pieces of potatoes in water. Mash the potato.
            Add some butter and salt and stir. Gradually add milk while stirring the potatoes. \\
        \end{minipage}
        &
        \begin{minipage}[t]{0.48\textwidth}
            \textbf{MART}: Cut the salmon into thin slices. \rept{Cut the salmon into thin slices}.
            \rept{Cut the salmon into thin slices}. \rept{Cut the salmon into thin slices}. \\
            \textbf{COOT (Ours)}: Cut the salmon in half. \rept{Cut the salmon in half}.
            \rept{Cut the salmon} into thin slices. \rept{Cut the salmon into thin slices}. \\
            \textbf{GT}: Slice the fish into smaller pieces. Chop the tail end off.
            Cut the fish at an angle. Cut the fish into thin pieces. \\
        \end{minipage}
        \\
    \end{tabularx}

    \label{tab:cap_anettest}
\end{table}

    \section{Related Work}\label{sec:related-work}

\narrowparagraph{Image and Language.} Many self-supervised visual-language representation learning methods
have focused on
improving one representation with the help of the other~\cite{klein15,kiros2014unifying,karpathy2014deep,stylese2019,
WangLL15,plummer2017phrase,kim2019mule}. These approaches learn joint image-text embeddings or map images and
sentences into a common space.
Recently, there has been a surging interest in utilizing Transformers for image-text representation
learning~\cite{chen2019uniter,qi2020imagebert,zhou2019vlp,huang2020pixelbert,miyazawa2020lambert}.
ViLBERT~\cite{lu2019vilbert} and VisualBERT~\cite{li2019visualbert} pretrain a BERT-like architecture on an
image-text dataset and then transfer learned representations to different downstream tasks.
LXMERT~\cite{tan2019lxmert} uses additional pretraining tasks and an object-relationship component. In contrast
to~\cite{tan2019lxmert,lu2019vilbert}, VL-BERT~\cite{su2019vl} does not utilize the task of sentence-image
relationship prediction and additionally pretrains the model on text-only datasets.

\narrowparagraph{Video and Language.}
The multi-modal nature of video is a great source of self-supervision. Modalities such as audio, text and motion
provide strong cues to learn richer spatio-temporal features~\cite{cvpr2019dong,miech18learning,miech19howto100m,
miech20endtoend,videotextretriv_1,PanMYLR15,sun2019videobert}.
Aytar et al.~\cite{AytarVT17} leverage natural synchronization to learn rich representations across vision, sound,
and language. VideoBERT~\cite{sun2019videobert} learns joint video-text representations based on predicting whether
the linguistic sentence is temporally aligned with the visual sentence.
These approaches~\cite{actbert20,sun2019contrastive,sun2019videobert} focus on self-supervised pretraining and
require a large set of paired video clips and texts to learn a good representation model~\cite{miech20endtoend}.


There has been growing interest in temporal localization of natural language in videos~\cite{zhang2018man,mun2020LGI,
hendricks17iccv,tall17,chenlook20}. Moment localization identifies a time window given a text
query~\cite{zhang2018man,mun2020LGI,gao2019wslln}.
Most related to our work are methods that focus on joint video-text embeddings and perform video-text retrieval or
captioning~\cite{gad13,gabeur2020multimodal,jmdvc15,cmhse,videotextretriv_1,PanMYLR15,venugopalan2015sequence,
chen2020finegrained,Shao_2018_ECCV}.
Several works tried to utilize the temporal structure of video and text for the alignment task~\cite{bojanowski15,
AytarVT17,book2movie,gad13,Yu_2018_ECCV}. Miech et al.~\cite{miech18learning} proposed a mixture of experts approach
to learn text-video representations.
Likewise, CE~\cite{Liu2019a} proposes a mixture-of-experts model to aggregate information from pretrained experts (e
.g.\ object, action, audio) with a gating mechanism.
In our work, we use a similar hierarchy as CMHSE~\cite{cmhse,hierarchical_2,hierarchical_3} and extend their design
by proposing three new components to learn the interactions between different levels of the hierarchy.


\narrowparagraph{Cycle-Consistency.} Cycle-Consistency uses transitivity as an objective for
training~\cite{cycle_consistency1,
CVPR2019_CycleTime,shah2019cycle,zhu2017unpaired}.
The assumption of cyclical structure has been used in various works~\cite{chech_cycle,cyclenips16,tang18}. Wang et al
.~\cite{CVPR2019_CycleTime} obtain the supervision for visual correspondence by tracking forward and backward.
Shah et al.~\cite{shah2019cycle} enforce consistency between the generated and the original question in visual
question answering. To prevent mode collapse, cycle-consistency is activated only after a certain number of training
iterations~\cite{shah2019cycle}. TCC~\cite{cycle_consistency1} employs cycle-consistency for temporal video alignment.
In contrast to TCC, which works only in the video domain, we align video and text.
To the best of our knowledge, this is the first work which introduces cycle-consistency to the video-text domain.

    \vspace{-3mm}
\section{Conclusions}\label{sec:conclusions}
\vspace{-2mm}
We have presented a cooperative hierarchical transformer architecture for learning a joint video and text embedding
space where similar semantics are aligned. The architecture is designed to encourage the use of long-range temporal
context in a cross-level manner.
Our approach uses two new components to model the interactions within and between hierarchy levels; an
attention-aware feature aggregation module to model the interactions between frames and words, a contextual
transformer to model the interactions between local contexts and global context.
In addition, we have introduced a new cross-modal cycle-consistency loss which enforces the semantic alignment of
clips and sentences.
We have shown that both components contribute -- jointly and individually -- to an improved retrieval performance. As
a result, our approach achieves state-of-the-art retrieval and captioning performance on two challenging datasets.


    \section*{Broader Impact}\label{sec:impact}

This work contributes fundamental research and does not present any
foreseeable societal consequence. In the long run, this line of research can
contribute to services on video search and video organisation.

    \begin{ack}
    We thank Ehsan Adeli for helpful comments, Antoine Miech for providing details on their retrieval evaluation,
    and Facebook for providing us a GPU server with Tesla P100 processors for this research work.
    \end{ack}

    \medskip

    {\small
        \bibliographystyle{unsrt}
        \bibliography{ref}
    }

    \renewcommand{\thesection}{\Alph{section}} 
    \setcounter{section}{0} 
    \renewcommand*{\theHsection}{A\thesection} 
    \clearpage
    \section{Appendix}
    


    \subsection{Implementation Details}\label{subsec:implementation-details}

\narrowparagraph{Hyperparameters.}
To select hyperparameters for our model, we used a combination of manual search and BOHB~\cite{bohb18} to explore the
hyperparameters space.
In Table~\ref{tab:hyperparams}, we provide an overview of our hyperparameter search.

After testing different activation functions ReLU, SELU~\cite{selu17}, ELU~\cite{elu16} and GELU~\cite{gelu1} we
found GELU to perform best.
Increasing the capacity of the model by using a higher attention dimension can help improve the results,
but comes at the cost of higher memory requirements and more difficult optimization.

\narrowparagraph{Optimization.}
\begin{table}[h]
    \caption{
        \textbf{Hyperparameters.} This table shows the hyperparameter ranges we considered and the final choices for our three best models
        (ActivityNet-captions, YouCook2-Resnet/Resnext features, Youcook2-Howto100m features.).
        ROP denotes the \textit{Reduce on Plateau} Scheduler we used.
        Dimensions given in multiples (1x, 2x) refer to multiples of the Attention Dimension parameter. FF denotes Feed-Forward.
        AF is our Attention-aware Feature Aggregation module.
    }
    \centering
    \small
    \resizebox{1\columnwidth}{!}{
    \begin{tabular}{l|rr|rrr}
        \thickhline
        \mr{Hyperparameter}                 & \multicolumn{2}{c|}{\mr{Considered Range}}    & \mr{ActivityNet}  & \multicolumn{2}{c}{Youcook2}  \\
                                            &   &                                           &                   & Resnet/ResneXt    & Howto100m \\ \thickhline
        Optimizer                           & \multicolumn{2}{c|}{Adam, RAdam, SGD}         & Adam              & RAdam             & RAdam     \\
        Learning rate                       & 1e-5  & 1e-2                                  & 1e-3              & 3.6e-4            & 9e-4      \\
        Weight Decay                        & 0     & 1e-2                                  & 2e-5              & 2e-5              & 0         \\
        Momentum                            & 0.5   & 0.99                                  & 0.9               & 0.56              & 0.56      \\
        Adam Beta2                          & 0.9   & 0.9999                                & 0.999             & 0.98              & 0.98      \\
        Adam Epsilon                        & 1e-10 & 1e-7                                  & 1e-8              & 1.5e-9            & 1.5e-9    \\
        Warmup Epochs                       & 0     & 8                                     & 3                 & 0                 & 0         \\
        ROP Patience                        & 2     & 10                                    & 2                 & 5                 & 5         \\
        ROP Cooldown                        & 0     & 3                                     & 3                 & 3                 & 3         \\ \hline
        Attention Layers                    & 1     & 3                                     & 1                 & 1                 & 1         \\
        Attention Dimension                 & 256   & 1024                                  & 384               & 384               & 384       \\
        Attention Heads                     & 1     & 8                                     & 8                 & 8                 & 8         \\
        Attention FF Dimension              & 1x    & 2x                                    & 1x                & 1x                & 1x        \\
        AF Dimension                        & 1x    & 2x                                    & 2x                & 2x                & 2x        \\
        AF Heads                            & 1     & 8                                     & 2                 & 2                 & 2         \\
        Number of AF modules                & 1     & 2                                     & 1                 & 1                 & 1         \\
        Dropout                             & 0\%   & 10\%                                  & 2.5\%             & 1\%               & 5\%       \\
        Gaussian Noise on Frame Features    & 0     & 1                                     & 0                 & 0.01              & 0         \\

\thickhline
\end{tabular}
}

\label{tab:hyperparams}
\end{table}

We tried several optimizers such as Adam, RAdam~\cite{liu2019radam} and SGD\@.
If carefully configured, RAdam can improve over Adam.

We schedule the Learning Rate with a \textit{Reduce on Plateau} approach:
Whenever our validation metric does not improve for a certain number of epochs, we reduce the learning rate by a
factor of 10.
After no improvements for 15 epochs, we terminate the training process.
As relevant metric we defined the sum of R@1 Retrieval Score for video-paragraph and paragraph-video retrieval on
Activitynet-Captions
and the sum of R@1 Retrieval Score for clip-sentence and sentence-clip retrieval on Youcook2.
Careful tuning of the optimizer parameters, using an automated search method like BOHB~\cite{bohb18} to search parts
of the parameters space,
was crucial to train the models properly.

\narrowparagraph{Strength of the cross-modal cycle-consistency loss.} For Activitynet we set $\lambda = 0.01$
and for Youcook2 we set $\lambda = 0.001$.

For weight initialization, we utilized Uniform, Normal and Truncated Normal distributions.
The best results were obtained with initializing weights randomly from the Truncated Normal distribution with a
standard deviation of $0.01$,
redrawing all samples with more than 2 standard deviations.

To cope with the overfitting problem, the different regularization methods (Dropout, Weight Decay, CMC-loss, Gaussian
Noise on Frame Features)
need to be traded off carefully to obtain good results (see Table~\ref{tab:hyperparams}).


\narrowparagraph{Preprocessing.}
For ActivityNet captions, we found it helpful to expand all clips to be at least 10 frames long.
Expanding is done by iteratively adding frames to the start and end of the clip until we reach the desired length.

\narrowparagraph{Retrieval.}
We L2-normalize the output embeddings of our model so the squared elements sum to 1.
Retrieval is done by cosine similarity, e.g.\ given video embedding $v$, we retrieve paragraph embedding
\begin{equation}
    p = \max_{\hat{p} \in \mathcal{D}}{v^\top \hat{p}}
    \label{eq:equation_retrieval}
\end{equation}


    \subsection{Ablation Studies}\label{subsec:experiments}
In this section, we provide ablation studies on the importance of low-level supervision, different text encoders 
performance,
impact of different alignment losses in our final training loss and analysis on sequence pooling.

\begin{table}[t]
    \caption{
        \textbf{Text feature ablation study on ActivityNet-captions (val1).}
        We evaluate our choice of text encoding and show that Bert~\cite{devlin2018bert} outperforms
        GloVe~\cite{pennington2014glove} on both models and all metrics.
    }
    \centering
    \small
    \setlength\tabcolsep{0.2cm}
    \renewcommand*{\arraystretch}{\arraystretchqual}
    \begin{tabular}{cccccccc}
        \thickhline
        \mr{Model} & \mr{Text} & \multicolumn{3}{c}{Paragraph$\implies$Video} & \multicolumn{3}{c}{Video$\implies$Paragraph} \\
        \cmidrule(rl){3-5}  \cmidrule(rl){6-8}
                &       & R@1   & R@5 & R@50 & R@1 & R@5 & R@50 \\ \thickhline
        HSE     & GloVe & \nsn{45.7}{0.3}   & \nsn{76.1}{0.7}   & \nsn{96.0}{0.3}   & \nsn{44.9}{0.5}   & \nsn{75.8}{1.2}   & \nsn{95.8}{0.4}   \\
        HSE     & Bert  & \nsn{47.0}{1.1}   & \nsn{77.0}{1.5}   & \nsn{96.1}{0.4}   & \nsn{46.9}{0.8}   & \nsn{77.2}{1.1}   & \nsn{95.9}{0.6}   \\ \hline
        \coot   & GloVe & \nsn{56.5}{1.1}   & \nsn{84.1}{1.3}   & \nsn{98.0}{0.3}   & \nsn{57.3}{1.8}   & \nsn{84.5}{1.4}   & \nsn{98.2}{0.2}   \\
        \coot   & Bert  & \nsb{60.8}{0.6}   & \nsb{86.6}{0.4}   & \nsb{98.6}{0.1}   & \nsb{60.9}{0.3}   & \nsb{87.4}{0.5}   & \nsb{98.6}{0.0}   \\
        \thickhline
    \end{tabular}
    \label{tab:ab_text}
\end{table}



\newcounter{magicrownumbers}
\newcommand\rownumber{\stepcounter{magicrownumbers}\arabic{magicrownumbers}}

\begin{table}[t]
    \caption{
        \textbf{Loss function ablation study on ActivityNet-captions (val1).}
        We analyse performance of the \coot \ model while removing loss components with different base models.
        CoT denotes using global attention in the contextual transformer. AF is our Attention-aware Feature Aggregation module.
    }
    \centering
    \small
    \setlength\tabcolsep{\tabcolsepqual}
    \begin{tabular*}{\linewidth}{@{\extracolsep{\fill}} r c cc ccccc ll ll}
        \thickhline
        \mr{\#} & {Pooling} & \mr{CMC} & \mr{CoT} & \multicolumn{3}{c}{Alignment}
        & \multicolumn{2}{c}{Clustering} & \multicolumn{2}{c}{Par.$\implies$Video} & \multicolumn{2}{c}{Video$\implies$Par.} \\
        \cmidrule(rl){2-2} \cmidrule(rl){5-7}  \cmidrule(rl){8-9} \cmidrule(rl){10-11} \cmidrule(rl){12-13}
        & Lowlvl & & & High & Low & Ctx & High & Low  & R@1 & R@5 & R@1 & R@5 \\ \thickhline
        \rownumber & Avg & \xmark & \xmark & \cmark & \xmark & \xmark & \xmark & \xmark  & \nsn{30.4}{3.2} & \nsn{58.4}{4.5} & \nsn{29.9}{3.3} & \nsn{58.7}{4.5} \\
        \rownumber & Avg & \xmark & \xmark & \cmark & \xmark & \cmark & \cmark & \xmark  & \nsn{49.7}{0.7} & \nsn{79.0}{0.6} & \nsn{48.6}{0.5} & \nsn{79.1}{0.9} \\
        \rownumber & Avg & \xmark & \xmark & \cmark & \xmark & \cmark & \cmark & \cmark  & \nsn{49.2}{0.7} & \nsn{78.9}{0.2} & \nsn{48.6}{0.6} & \nsn{78.9}{0.6} \\
        \rownumber & Avg & \xmark & \xmark & \cmark & \cmark & \xmark & \cmark & \cmark  & \nsn{50.6}{1.1} & \nsn{79.8}{0.8} & \nsn{50.8}{1.0} & \nsn{79.8}{0.8} \\
        \rownumber & Avg & \xmark & \xmark & \cmark & \cmark & \cmark & \xmark & \xmark  & \nsn{51.5}{0.7} & \nsn{80.2}{0.4} & \nsn{52.0}{0.8} & \nsn{80.5}{0.3} \\
        \rownumber & Avg & \xmark & \xmark & \cmark & \cmark & \cmark & \cmark & \cmark  & \nsn{52.6}{0.6} & \nsn{80.6}{0.4} & \nsn{52.1}{0.4} & \nsn{80.8}{0.2} \\
        \hline
        \rownumber & Avg & \cmark & \xmark & \cmark & \xmark & \xmark & \xmark & \xmark  & \nsn{27.4}{2.1} & \nsn{55.3}{2.4} & \nsn{27.3}{1.6} & \nsn{56.0}{2.3} \\
        \rownumber & Avg & \cmark & \xmark & \cmark & \cmark & \cmark & \cmark & \cmark  & \nsn{54.1}{0.8} & \nsn{82.0}{0.1} & \nsn{54.7}{0.2} & \nsn{82.1}{0.1} \\
        \rownumber & Avg & \cmark & \cmark & \cmark & \cmark & \cmark & \cmark & \cmark  & \nsn{53.6}{0.1} & \nsn{81.7}{0.0} & \nsn{53.5}{0.5} & \nsn{81.7}{0.7} \\
        \hline
        \rownumber & Max & \xmark & \xmark & \cmark & \xmark & \xmark & \xmark & \xmark  & \nsn{47.9}{0.7} & \nsn{76.9}{0.1} & \nsn{48.3}{0.2} & \nsn{77.5}{0.6} \\
        \rownumber & Max & \xmark & \xmark & \cmark & \xmark & \cmark & \cmark & \xmark  & \nsn{56.5}{0.3} & \nsn{84.5}{0.2} & \nsn{56.6}{0.4} & \nsn{85.2}{0.1} \\
        \rownumber & Max & \xmark & \xmark & \cmark & \xmark & \cmark & \cmark & \cmark  & \nsn{54.4}{0.9} & \nsn{83.3}{0.9} & \nsn{55.4}{1.4} & \nsn{84.0}{0.8} \\
        \rownumber & Max & \xmark & \xmark & \cmark & \cmark & \xmark & \cmark & \cmark  & \nsn{55.3}{0.8} & \nsn{83.0}{0.8} & \nsn{56.4}{1.1} & \nsn{83.7}{1.2} \\
        \rownumber & Max & \xmark & \xmark & \cmark & \cmark & \cmark & \xmark & \xmark  & \nsn{56.1}{0.2} & \nsn{83.3}{0.2} & \nsn{57.0}{0.3} & \nsn{83.9}{0.5} \\
        \rownumber & Max & \xmark & \xmark & \cmark & \cmark & \cmark & \cmark & \cmark  & \nsn{58.2}{0.5} & \nsn{84.9}{0.2} & \nsn{58.7}{0.5} & \nsn{86.0}{0.2} \\
        \hline
        \rownumber & Max & \cmark & \xmark & \cmark & \xmark & \xmark & \xmark & \xmark  & \nsn{46.3}{1.0} & \nsn{76.2}{0.9} & \nsn{47.7}{0.9} & \nsn{77.2}{0.7} \\
        \rownumber & Max & \cmark & \xmark & \cmark & \cmark & \cmark & \cmark & \cmark  & \nsn{57.5}{0.5} & \nsn{84.8}{0.2} & \nsn{58.1}{1.0} & \nsn{85.3}{0.4} \\
        \rownumber & Max & \cmark & \cmark & \cmark & \cmark & \cmark & \cmark & \cmark  & \nsn{59.4}{0.9} & \nsn{86.1}{0.6} & \nsn{60.5}{1.0} & \nsn{87.1}{0.2} \\
        \hline
        \rownumber & AF  & \xmark & \xmark & \cmark & \xmark & \xmark & \xmark & \xmark  & \nsn{47.1}{0.7} & \nsn{76.7}{0.6} & \nsn{47.6}{0.2} & \nsn{77.4}{0.2} \\
        \rownumber & AF  & \xmark & \xmark & \cmark & \xmark & \cmark & \cmark & \xmark  & \nsn{56.3}{0.3} & \nsn{84.0}{0.2} & \nsn{56.8}{0.7} & \nsn{84.7}{0.3} \\
        \rownumber & AF  & \xmark & \xmark & \cmark & \xmark & \cmark & \cmark & \cmark  & \nsn{55.2}{0.2} & \nsn{83.3}{0.1} & \nsn{55.8}{0.5} & \nsn{83.6}{0.2} \\
        \rownumber & AF  & \xmark & \xmark & \cmark & \cmark & \xmark & \cmark & \cmark  & \nsn{57.8}{0.3} & \nsn{84.6}{0.2} & \nsn{58.1}{0.3} & \nsn{85.1}{0.2} \\
        \rownumber & AF  & \xmark & \xmark & \cmark & \cmark & \cmark & \xmark & \xmark  & \nsn{58.8}{0.4} & \nsn{85.3}{0.4} & \nsn{59.1}{0.6} & \nsn{85.8}{0.4} \\
        \rownumber & AF  & \xmark & \xmark & \cmark & \cmark & \cmark & \cmark & \cmark  & \nsn{59.0}{0.5} & \nsn{85.4}{0.2} & \nsn{59.8}{0.6} & \nsn{85.8}{0.8} \\
        \hline
        \rownumber & AF  & \cmark & \xmark & \cmark & \xmark & \xmark & \xmark & \xmark  & \nsn{47.8}{0.5} & \nsn{76.4}{0.4} & \nsn{47.8}{0.2} & \nsn{77.5}{0.5} \\
        \rownumber & AF  & \cmark & \xmark & \cmark & \cmark & \cmark & \cmark & \cmark  & \nsn{59.5}{0.5} & \nsn{85.5}{0.4} & \nsn{60.5}{0.7} & \nsn{86.2}{0.5} \\
        \rownumber & AF  & \cmark & \cmark & \cmark & \xmark & \xmark & \xmark & \xmark  & \nsn{53.9}{0.7} & \nsn{82.6}{0.6} & \nsn{53.8}{0.6} & \nsn{83.0}{0.5} \\
        \rownumber & AF  & \cmark & \cmark & \cmark & \cmark & \cmark & \xmark & \xmark  & \nsn{55.1}{5.3} & \nsn{83.4}{3.6} & \nsn{55.5}{4.7} & \nsn{83.8}{3.2} \\
        \rownumber & AF  & \cmark & \cmark & \cmark & \cmark & \xmark & \cmark & \cmark  & \nsn{58.5}{1.1} & \nsn{85.2}{0.5} & \nsn{58.5}{0.7} & \nsn{85.5}{0.7} \\
        \hline
        \rownumber & AF  & \cmark & \cmark & \cmark & \cmark & \cmark & \cmark & \cmark & \nsb{60.8}{0.6} & \nsb{86.6}{0.4} & \nsb{60.9}{0.3} & \nsb{87.4}{0.5} \\
        \thickhline
    \end{tabular*}
    \label{tab:ab_loss}
\end{table}

\begin{table}[t]
    \caption{
        \textbf{Evaluation of different sequence pooling methods on ActivityNet-captions (val1).}
        We switch both the low level (frames, words) and high level (clips, sentences) pooling methods and
        observe the changes in performance.
        In experiments denoted with *, we used a different optimizer setting.
    }
    \centering
    \small
    \begin{tabular*}{\textwidth}{@{\extracolsep{\fill}}cccccccccc}
        \thickhline
        \multicolumn{2}{c}{Pooling} & \mr{CMC} & \mr{CoT} & \multicolumn{2}{c}{Par.$\implies$Video} & \multicolumn{2}{c}{Video$\implies$Par.} \\
        \cmidrule(rl){1-2}                                  \cmidrule(rl){5-6}                        \cmidrule(rl){7-8}
        Low     & High  &           &           & R@1               & R@5               & R@1               & R@5               \\ \thickhline
        \hline
        AF      & AF x1 & \cmark    & \cmark    & \nsn{42.6}{0.4}   & \nsn{76.5}{0.3}   & \nsn{42.1}{0.8}   & \nsn{76.9}{0.7}   \\
        AF      & AF x2 & \cmark    & \cmark    & \nsn{42.8}{0.1}   & \nsn{76.0}{0.7}   & \nsn{42.8}{0.1}   & \nsn{76.4}{0.6}   \\
        AF*     & AF x1 & \cmark    & \cmark    & \nsn{48.7}{1.0}   & \nsn{82.2}{0.6}   & \nsn{50.1}{0.4}   & \nsn{82.6}{0.4}   \\
        AF*     & AF x2 & \cmark    & \cmark    & \nsn{50.5}{0.4}   & \nsn{82.3}{0.4}   & \nsn{51.4}{1.4}   & \nsn{82.9}{0.6}   \\ \hline
        Max     & Max   & \cmark    & \cmark    & \nsn{40.9}{0.7}   & \nsn{75.3}{0.1}   & \nsn{42.2}{0.5}   & \nsn{76.2}{0.6}   \\
        AF      & Max   & \cmark    & \cmark    & \nsn{43.3}{0.9}   & \nsn{76.3}{1.0}   & \nsn{42.5}{0.6}   & \nsn{77.2}{1.2}   \\ \hline
        CLS     & Avg   & \xmark    & \xmark    & \nsn{49.4}{1.4}   & \nsn{77.7}{1.3}   & \nsn{49.7}{1.9}   & \nsn{77.8}{0.9}   \\
        CLS     & Avg   & \cmark    & \cmark    & \nsn{49.7}{0.5}   & \nsn{79.4}{0.2}   & \nsn{51.2}{0.1}   & \nsn{79.6}{0.1}   \\
        Avg     & Avg   & \xmark    & \xmark    & \nsn{52.6}{0.6}   & \nsn{80.6}{0.4}   & \nsn{52.1}{0.4}   & \nsn{80.8}{0.2}   \\
        Avg     & Avg   & \cmark    & \cmark    & \nsn{53.6}{0.1}   & \nsn{81.7}{0.0}   & \nsn{53.5}{0.5}   & \nsn{81.7}{0.7}   \\
        Max     & Avg   & \xmark    & \xmark    & \nsn{58.2}{0.5}   & \nsn{84.9}{0.2}   & \nsn{58.7}{0.5}   & \nsn{86.0}{0.2}   \\
        Max     & Avg   & \cmark    & \cmark    & \nsn{59.4}{0.9}   & \nsn{86.1}{0.6}   & \nsn{60.5}{1.0}   & \nsn{87.1}{0.2}   \\ \hline
        AF      & Avg   & \cmark    & \cmark    & \nsb{60.8}{0.6}   & \nsb{86.6}{0.4}   & \nsb{60.9}{0.3}   & \nsb{87.4}{0.5}   \\
        \thickhline
    \end{tabular*}
    \label{tab:ab_pooling}
\end{table}

\begin{table}[t]
    \caption{
        \textbf{Evaluation of different averagepooling methods.}
        We modify our exact approach to averagepooling in the high-level and evaluate the results.
    }
    \centering
    \small
    \setlength\tabcolsep{\tabcolsepquant}
    \renewcommand*{\arraystretch}{\arraystretchquant}

    \vspace{0.2cm} \textbf{ActivityNet-captions} dataset: \\ \vspace{0.2cm}
    \begin{tabular*}{\textwidth}{@{\extracolsep{\fill}}ccccccc}
        \thickhline
        \mr{Sum}    & \mr{Pad}          & \mr{Divide}   & \multicolumn{2}{c}{Par.$\implies$Video}   & \multicolumn{2}{c}{Video$\implies$Par.}   \\
                                                          \cmidrule(rl){4-5}                          \cmidrule(rl){6-7}
                    &                   &               & R@1               & R@5                   & R@1               & R@5                   \\ \thickhline
        All         & Max(Batch, 16)    & All           & \nsn{44.2}{2.5}   & \nsn{75.4}{2.5}       & \nsn{44.1}{2.0}   & \nsn{75.9}{2.0}       \\
        All         & Max(Batch, 16)    & Nonzero       & \nsn{44.1}{0.7}   & \nsn{76.2}{0.9}       & \nsn{44.7}{1.1}   & \nsn{76.9}{0.9}       \\
        All         & Batch             & All           & \nsn{48.3}{0.2}   & \nsn{76.8}{0.8}       & \nsn{47.9}{0.8}   & \nsn{77.7}{0.7}       \\
        Nonzero     & Batch             & Nonzero       & \nsn{42.0}{0.5}   & \nsn{76.3}{0.3}       & \nsn{41.6}{0.6}   & \nsn{77.0}{0.7}       \\ \hline
        All         & Batch             & Nonzero       & \nsb{60.8}{0.6}   & \nsb{86.6}{0.4}       & \nsb{60.9}{0.3}   & \nsb{87.4}{0.5}       \\ \thickhline
    \end{tabular*}

    \vspace{0.2cm} \textbf{Youcook2} dataset:
    \\ \vspace{0.2cm}
    \begin{tabular*}{\textwidth}{@{\extracolsep{\fill}}ccccccc}
        \thickhline
        \mr{Sum}    & \mr{Pad} & \mr{Divide} & \multicolumn{2}{c}{Par.$\implies$Video} & \multicolumn{2}{c}{Sent.$\implies$Clip} \\
        \cmidrule(rl){4-5} \cmidrule(rl){6-7}
        & &                                 & R@1  & R@5  & R@1  & R@5  \\ \thickhline
        All      & Max(Batch, 16) & All     & \nsb{77.6}{0.7} & \nsb{96.3}{0.4} & \nsb{17.5}{0.3} & \nsb{40.7}{0.1} \\ 
        All      & Max(Batch, 16) & Nonzero & \nsn{74.7}{2.0} & \nsn{95.0}{0.6} & \nsn{16.9}{0.5} & \nsn{39.7}{0.8} \\ 
        All      & Batch          & All     & \nsn{77.4}{1.5} & \nsn{96.2}{1.6} & \nsn{17.2}{0.6} & \nsn{39.9}{0.3} \\ 
        Nonzero  & Batch          & Nonzero & \nsn{74.2}{2.6} & \nsn{94.7}{0.7} & \nsn{16.8}{0.3} & \nsn{40.2}{0.6} \\ \hline
        All      & Batch          & Nonzero & \nsn{77.2}{1.0}   & \nsn{95.8}{0.8}   & \nsn{16.7}{0.4}   & \nsn{40.2}{0.3}  \\

        \thickhline
    \end{tabular*}

    \label{tab:ab_avgpooling}
\end{table}




\begin{table}[t]
    \caption{
        \textbf{Video-paragraph retrieval results on AcitvityNet-captions dataset (val2).}
        }
    \centering
    \small
    \renewcommand*{\arraystretch}{1.1} 
    \begin{tabular*}{0.7\textwidth}{@{\extracolsep{\fill}}crrrr}
        \thickhline
        \mr{Method} & \multicolumn{2}{c}{Par.$\implies$Video} & \multicolumn{2}{c}{Video$\implies$Par.} \\
        \cmidrule(rl){2-3}\cmidrule(rl){4-5}
         & R@1               & R@5               & R@1               & R@5               \\ \thickhline
        \hline
        FSE & 11.5 & 31.0 & 11.0 & 30.6 \\
        HSE & 32.9 & 62.7 & 32.6 & 63.0 \\\hline
        COOT & \textbf{48.5} & \textbf{78.9} & \textbf{48.9} & \textbf{79.5} \\
        \thickhline
    \end{tabular*}
    \label{tab:retr_anetval2}
\end{table}

\begin{figure}
    \caption{Noise vs Performance study on ActivityNet-captions dataset (val1)}
    \centering
    \includegraphics[width=0.5\textwidth]{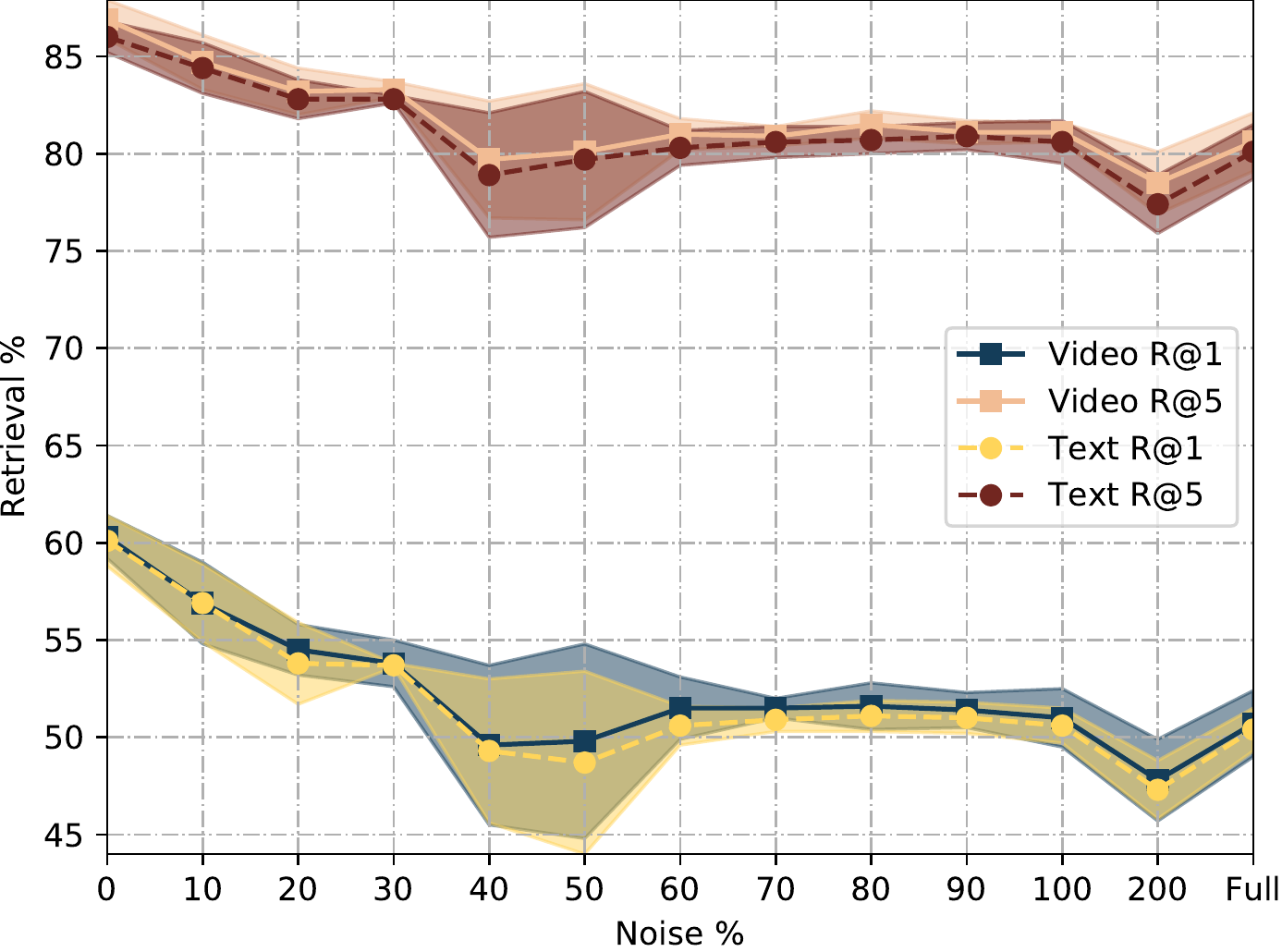}
    \label{fig:ab_noise}
\end{figure}

\narrowparagraph{Importance of low-level supervision.} In Fig.~\ref{fig:ab_noise}, we study the effect of adding uniform noise
to the start and end frame index of each clip in ActivityNet-captions from the interval $[-N_f*P, +N_f*P]$. $N_f$ is
the total number of video frames and $P$ is the noise percentage. We also perform a "full" noise experiment where we
drop the temporal alignment labels of clips and sentences completely.
We observe that increasing the noise from 0\% to 40\% consistently decreases the performance as labels get less
reliable. For noise more than 40\%, we do not observe significant changes in performance anymore.
This is probably because at this noise level the labels become useless and are ignored. Still a good performance is
obtained.


The study shows that the model is robust to noisy and missing low-level supervision and \coot \ is still able to
capture useful dynamics between low-level and high-level semantics.

\narrowparagraph{Impact of Text Encoding.}
We conduct ablation experiments to evaluate the importance of the text encoder for representation learning task.
The ablation study results are shown in Table~\ref{tab:ab_text}.
We first evaluate the COOT model and the HSE~\cite{cmhse} model with GloVe~\cite{pennington2014glove} features.
We then replace GloVe features with features obtained from a pretrained Bert~\cite{devlin2018bert} model.
Note that we feed an entire paragraph consisting of several sentences into Bert, leveraging high-level context.
Our results show that replacing fixed word embeddings with the context-aware Bert features can significantly improve 
model performance over different architectures.
Both models are relatively shallow (1 layer of attention / GRU respectively), which may be the reason why the deeper 
Bert model (13 layers) can help understand the text better.

\narrowparagraph{Impact of Alignment Losses.}
We study the effect of alignment losses on the performance in Table~\ref{tab:ab_loss}.
To give a more diverse picture, we evaluate the losses under different settings:
We use three different low level pooling methods (Averagepool, Maxpool and Attention-aware Feature Aggregation)
and selective disable the Cross-Modal Cycle Consistency loss and the global context attention in the Contextual 
Transformer.


We found that removing any or all of the three alignment losses significantly decreases performance.
In addition, we observed that clustering losses have a positive impact on the performance of the model.

Note that we also tried clustering the global context and found it to be not helpful. It might be a too strong 
constraint on our low-level embedding network.

\narrowparagraph{Study on sequence pooling methods.}
In Table~\ref{tab:ab_pooling} we replace our low-level (frames, words) and high-level (clips, sentences) pooling 
methods and evaluate the performance.
Interestingly, we get the best results with our AF module on the low level, while averagepooling outperforms it on 
the high level.
Removing our components (CMC, CoT) and replacing AF with maxpooling provides a considerably strong baseline compared 
to our full model.

The sequence length is higher on the low level (e.g.\ up to 80 frames) than on the high level (on average 3.6 clips 
per video).
Additionally, there are stronger temporal relationships between semantics in the low level. The AF module can learn 
to capture these relationships and improve the features.
However on the high-level, the semantics have more independent meanings which makes it much harder for AF to model 
the temporal relationships between them.

Note that to give a more fair comparison, we change the optimizer setting when adding AF on the high level, as 
denoted with *.
We observe that concatenating the output of 2 AF modules on the high level improves the performance, suggesting that 
the two modules learn to attend to different features.

We also vary our approach on averagepooling on the high level and report results In Table~\ref{tab:ab_avgpooling}.
Working on variable length inputs, there are a number of design choices to make. We evaluate the following ones:
\emph{1)} Summing over the unmasked sequence elements (nonzero inputs) only or summing over both sequence and
padding elements (zero inputs).
\emph{2)} Minimum padding to the maximum sequence length in the minibatch or to a length of at least 16.
\emph{3)} Obtaining the average by dividing the sum by the length of nonzero elements or by the length of all
elements.

On \textbf{ActivityNet-captions} (split val1, average sequence length 3.6),
we show our non-standard approach of including padding tokens in the sum but dividing by the length of non-padding 
tokens works well.
Note that in all other reported experiments, we use this version of averagepooling.

On \textbf{Youcook2} (split val, average sequence length 7.6), we cannot reproduce this large gain in performance but
the approach still works reasonably well.
The good results when padding to a minimum length of 16 might be due to the average length being closer to 16 than in
ActivityNet-captions.

\subsection{Retrieval on ActivityNet-captions (split val2)}\label{subsec:retrievalanet}

We provide retrieval results for ActivityNet-Captions (val2 split) in Table~\ref{tab:retr_anetval2}.


    \subsection{Qualitative Results}\label{sec:qualitative-results}
\begin{table}[t]
    \caption{
        \textbf{Qualitative Results on Activitynet for Paragraph-to-Video Retrieval.}
        For each text query, we show some frames from the top three ranked videos together with the correct video.
        For clarification, we show video results with text.
        \textbf{Left:} The correct video has a high rank and all top results are very relevant to the query.
        \textbf{Right:} Even though the correct video is ranked low, the top videos are semantically similar to the text query.
    }
    \centering
    \small
    \setlength\tabcolsep{\tabcolsepqual}
    \renewcommand*{\arraystretch}{\arraystretchqual}
    \begin{tabularx}{\linewidth}{l|X|l|X}
        \thickhline
        \multicolqual{X|}{
            \textit{Query:} A man is standing inside a workshop. He leans over, welding a piece of metal.
            Sparks fly as he welds. } &
        \multicolqual{X}{
            \textit{Query:} A person is kneeling down painting something on the ground. They smooth out the paint.
                They continue painting layers on top of the paint. } \\ \hline
        \rankscoretitle{Rank}{Score}    & Retrieved Video                   & \rankscoretitle{Rank}{Score}  & Retrieved Video                   \\ \hline
        \rankscore{1}{0.827}            & \imagequal{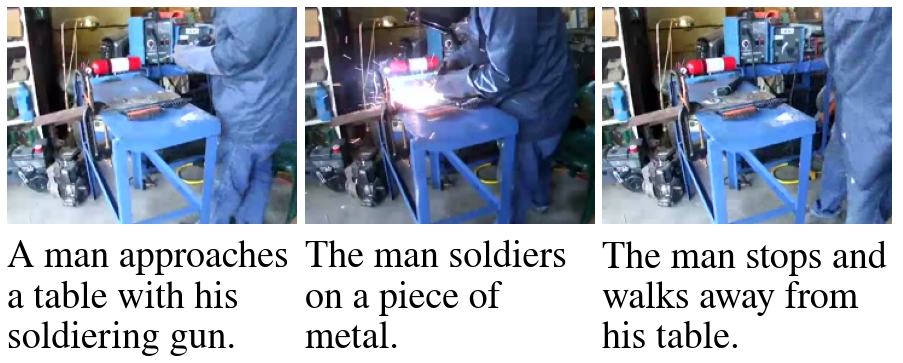}    & \rankscore{1}{0.654}          & \imagequal{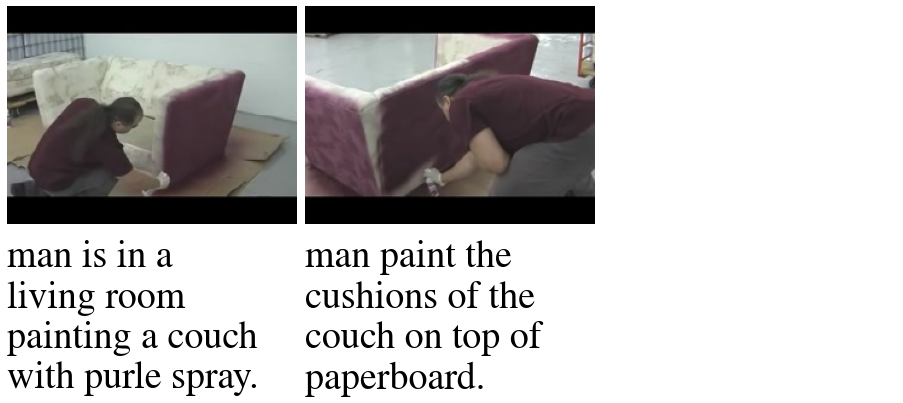}    \\ \hline
        \rankscoremark{2}{0.821}        & \imagequal{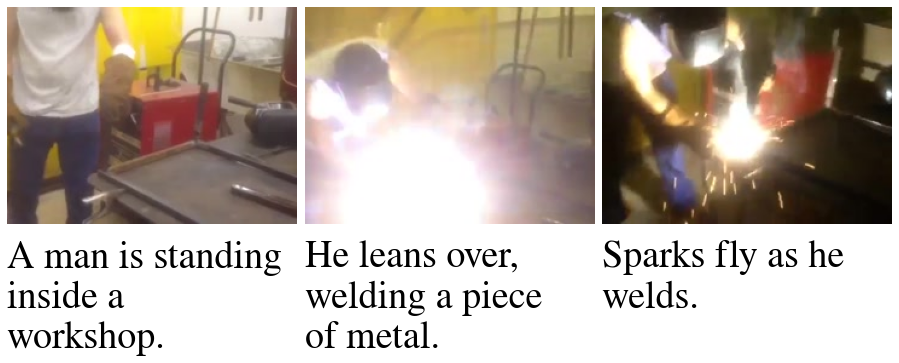}    & \rankscore{2}{0.643}          & \imagequal{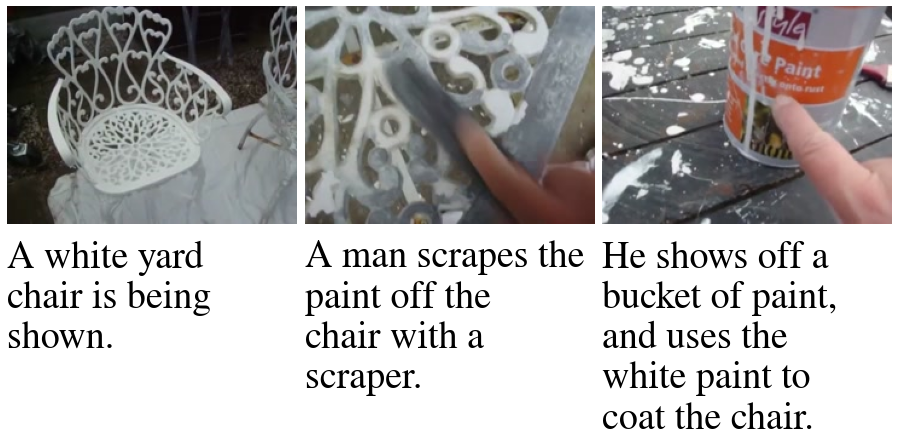}    \\ \hline
        \rankscore{3}{0.816}            & \imagequal{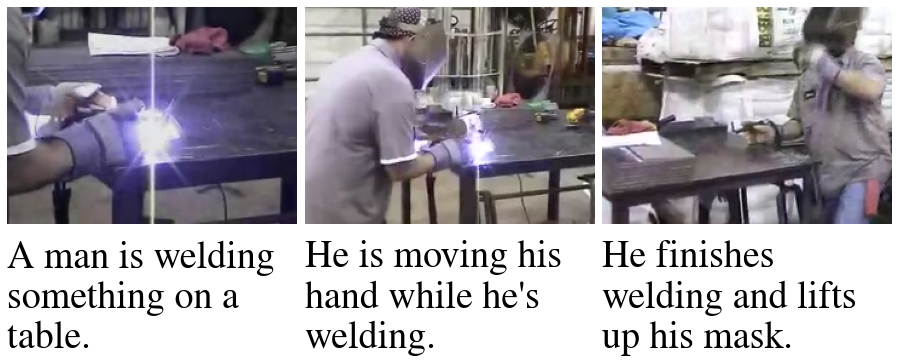}    & \rankscore{3}{0.640}          & \imagequal{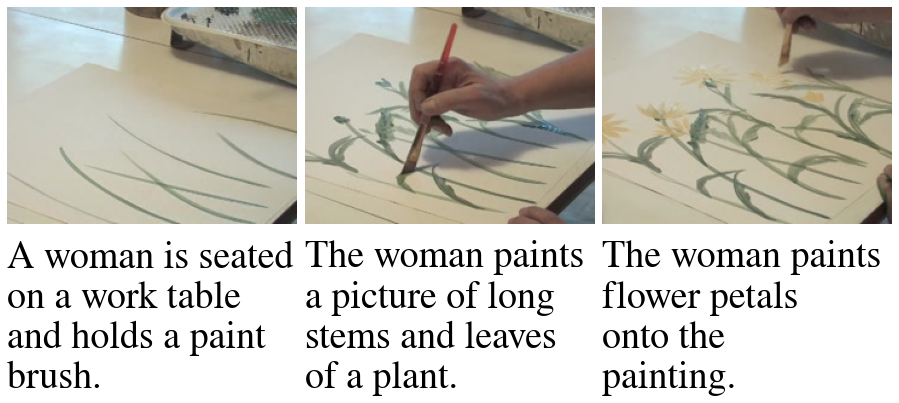}    \\ \hline
        \rankscore{4}{0.783}            & \imagequal{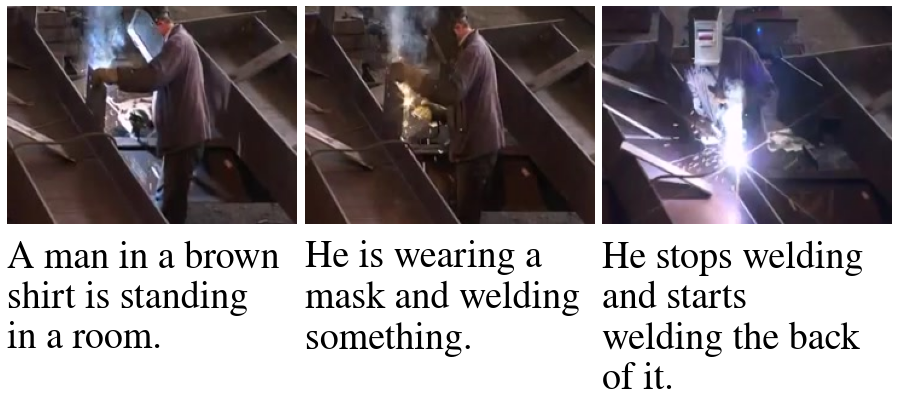}    & \rankscoremark{48}{0.438}     & \imagequal{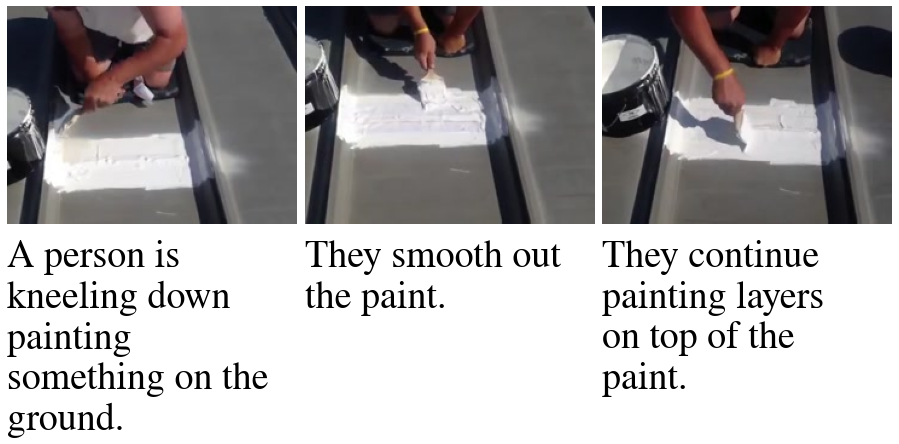}    \\ \thickhline
    \end{tabularx}

    \label{tab:retr_anet_p2v}
\end{table}

\begin{table}[t]
    \caption{
        \textbf{Retrieval Video to Paragraph on Activitynet.}
        Long paragraphs have been shortened, as indicated by "[...]".
        \textbf{Left:} The correct paragraph is identified with a considerable score margin to the 2nd place.
        \textbf{Right:} The top results are from the same activity as the input video (\textit{dancing}).
    }
    \centering
    \small
    \setlength\tabcolsep{\tabcolsepqual}
    \renewcommand*{\arraystretch}{\arraystretchqual}
    \begin{tabularx}{\linewidth}{l|X|l|X}
        \thickhline
        \multicolqual{X|}{\textit{Query:}} &
        \multicolqual{X} {\textit{Query:}} \\
        \multicolqual{X|}{\imagequeryqual{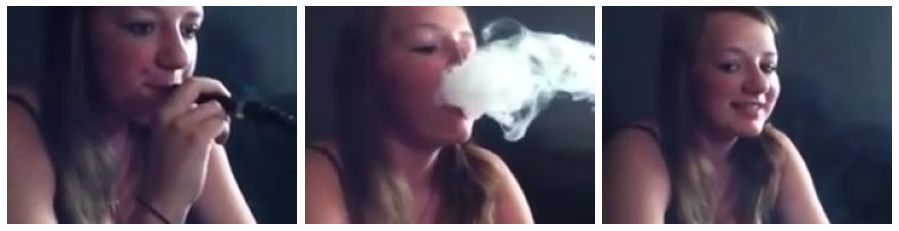}} &
        \multicolqual{X} {\imagequeryqual{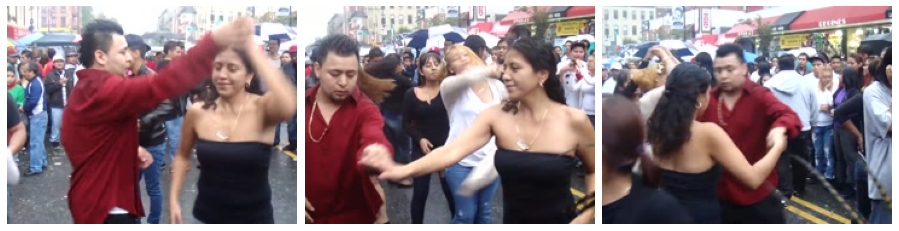}} \\ \hline
        \rankscoretitle{Rank}{Score}    & Retrieved Text & \rankscoretitle{Rank}{Score} & Retrieved Text \\ \hline
        \rankscoremark{1}{0.813}        &
        \textbf{A woman is resting next to crashing water. She is smoking a pipe. She blows out a plume of smoke.} &
        \rankscore{1}{0.717}            &
        A woman stands in front of a crowd of people on a public sidewalk and dances with a male dance partner
        in ballroom style dance. [\ldots] \\ \hline
        \rankscore{2}{0.654}            &
        A close up of a man's chin is shown followed by him smoking a hookah pipe. He takes the pipe out
        of his mouth and blows the smoke into the camera. &
        \rankscore{2}{0.678}            &
        A woman in a leather dress and hat dances in a public station. A man joins her, dancing side to side in
        a flamenco style dance. They continue dancing as a small crowd gathers to watch. [\ldots] \\ \hline
        \rankscore{3}{0.641}            &
        A close up of tin foil is shown leading a woman taking a large hit out of a hookah hose. She
        continues smoking out of the hookah [\ldots] &
        \rankscore{3}{0.608}            &
        A large group of people are seen standing around a city center waiting for people to arrive.
        Girls dancing are seen walking through the parade as other people watch on the side. [\ldots] \\ \hline
        \rankscore{4}{0.601}            &
        A woman is laying back in a chair getting her lip pierced. The piercer removes the tool and pulls on her lip. &
        \rankscoremark{16}{0.496}       &
        \textbf{People are dancing in a street. People are standing on the sidelines watching them.
        They continue dancing on a street.} \\ \thickhline
    \end{tabularx}
    \label{tab:retr_anet_v2p}
\end{table}



\begin{table}[t]
    \caption{
        \textbf{Sentence-to-Clip Retrieval on Youcook2.}
        For clarification, we show clip results with corresponding text.
        \textbf{Left:} The model ranks the correct video at the top and even distinguishes
        it from other videos about the same activity.
        \textbf{Right:} The \textit{slicing} task is correctly recognized, but the model is
        not able to understand which object is being chopped (\textit{bamboo shots}).}
    \centering
    \small
    \setlength\tabcolsep{\tabcolsepqual}
    \renewcommand*{\arraystretch}{\arraystretchqual}
    \begin{tabularx}{\linewidth}{l|X|l|X}
        \thickhline
        \multicolqual{X|}{\textit{Query:} melt butter in the pan } &
        \multicolqual{X} {\textit{Query:} slice the bamboo shoots into strips } \\ \hline
        \rankscoretitle{Rank}{Score}    & Retrieved Clip                    & \rankscoretitle{Rank}{Score}  & Retrieved Clip                 \\ \hline
        \rankscoremark{1}{0.642}        & \imagequal{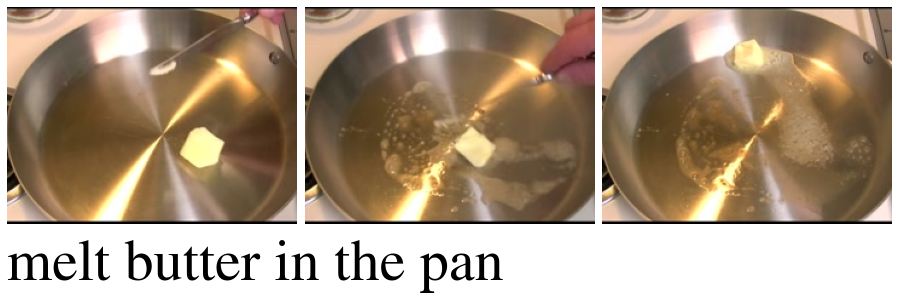}     & \rankscore{1}{0.621}          & \imagequal{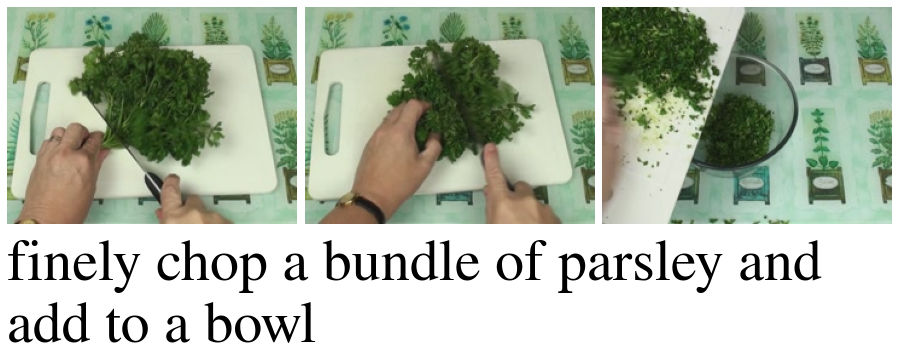}  \\ \hline
        \rankscore{2}{0.583}            & \imagequal{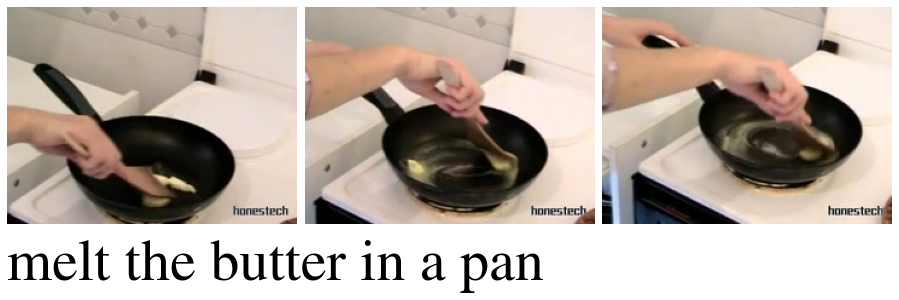}     & \rankscore{2}{0.610}          & \imagequal{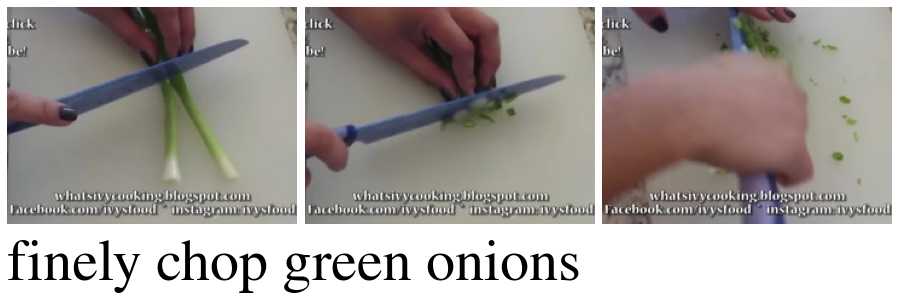}  \\ \hline
        \rankscore{3}{0.561}            & \imagequal{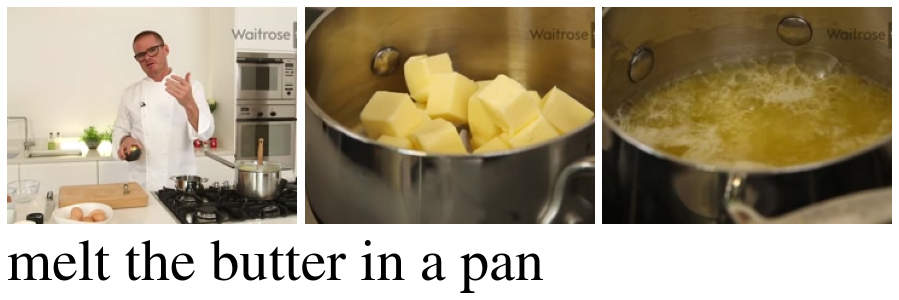}     & \rankscore{3}{0.609}          & \imagequal{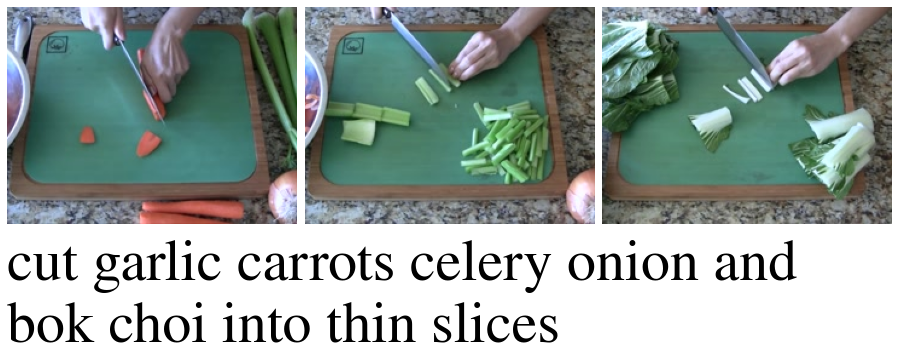}  \\ \hline
        \rankscore{4}{0.553}            & \imagequal{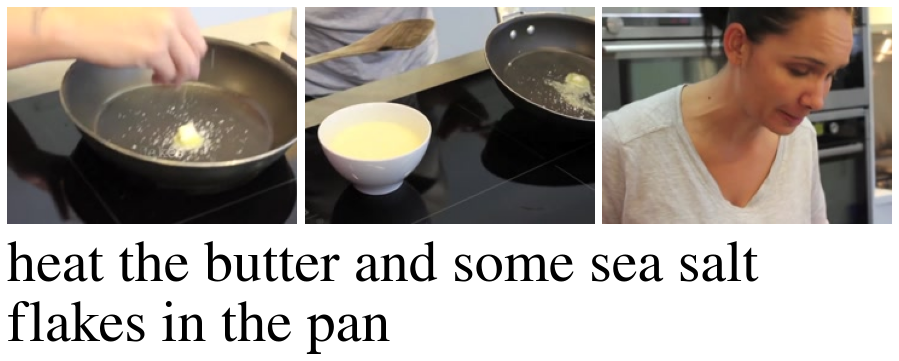}     & \rankscoremark{168}{0.326}    & \imagequal{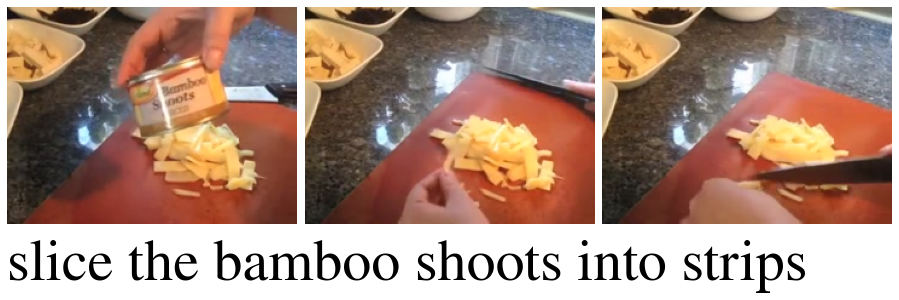}  \\ \thickhline
    \end{tabularx}
    \label{tab:retr_yk_p2v}
\end{table}

\begin{table}[t]
    \caption{\textbf{Clip-to-Sentence Retrieval on Youcook2 val set.}
    \textbf{Left:} The model gives high relative score to the relevant text but has problems visually
    distinguishing \textit{apples} from \textit{potatoes}. \textbf{Right:}: \textit{Wine} is confused
    with \textit{oil} and the herbs cannot be identified precisely to be \textit{bay leaves} and
    \textit{thyme}. Identical sentences can produce different results, since the Bert~\cite{devlin2018bert}
    text encoder takes paragraph context into account and therefore the model inputs differ.}
    \centering
    \small
    \setlength\tabcolsep{\tabcolsepqual}
    \renewcommand*{\arraystretch}{\arraystretchqual}
    \begin{tabularx}{\textwidth}{l|X|l|X}
        \thickhline
        \multicolqual{X|}{\textit{Query:}} &
        \multicolqual{X} {\textit{Query:}} \\
        \multicolqual{X|}{\imagequeryqual{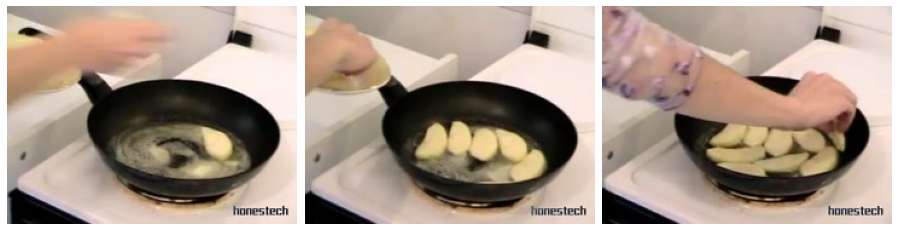}} &
        \multicolqual{X} {\imagequeryqual{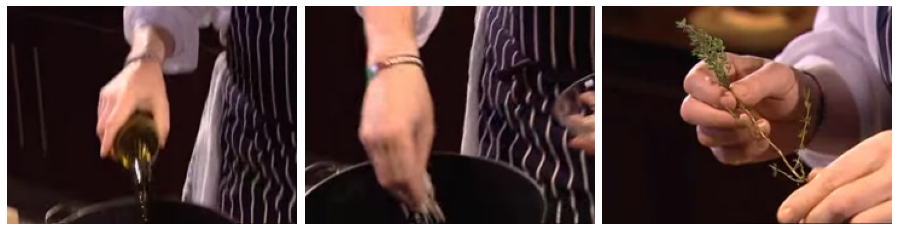}} \\ \hline
        \rankscoretitle{Rank}{Score}    & Retrieved Text &
        \rankscoretitle{Rank}{Score}    & Retrieved Text \\ \hline
        \rankscore{1}{0.523}            & place the potato wedges into a pan of hot oil &
        \rankscore{1}{0.705}            & add oil and herbs to a pan \\ \hline
        \rankscoremark{2}{0.514}        & \textbf{cook the apple slices in the pan} &
        \rankscore{2}{0.622}            & heat oil to 365 in a pan \\ \hline
        \rankscore{3}{0.510}            & remove the potatoes from the oil and place on paper towel &
        \rankscore{3}{0.603}            & heat some oil in a pan \\ \hline
        \rankscore{4}{0.497}            & add oil to the pan and fry the hash browns &
        \rankscore{4}{0.579}            & heat some oil in a pan \\ \hline
        \rankscore{5}{0.495}            & fry the potatos in oil &
        \rankscore{5}{0.575}            & add oil to a pan \\ \hline
        \rankscore{6}{0.480}            & add the potatoes to the pan &
        \rankscore{6}{0.570}            & heat some olive oil in a pan \\ \hline
        \rankscore{7}{0.477}            & heat the apple in a pan with some oil &
        \rankscore{7}{0.567}            & heat some oil in a pan \\ \hline
        \rankscore{8}{0.475}            & pierce the knife inside the potatoes and find if the potatoes are cooked properly &
        \rankscore{8}{0.564}            & heat oil in a pan \\ \hline
        \rankscore{9}{0.474}            & melt little butter and olive oil in a pan &
        \rankscore{9}{0.564}            & heat some oil cumin seeds and coriander seeds in a pan \\ \hline
        \rankscore{10}{0.470}           & fry the potatoes in a deep fryer &
        \rankscoremark{85}{0.385}       & \textbf{add white wine onions a bay leaf and thyme to the pot} \\ \hline
        \thickhline
    \end{tabularx}
    \label{tab:retr_yk_v2p}
\end{table}

\begin{table}[t]
    \caption{
    \textbf{Random Captioning samples on YouCook2 (val split).}
    }
    \centering
    \small
    \setlength\tabcolsep{\tabcolsepqual}
    \renewcommand*{\arraystretch}{\arraystretchqual}

    \begin{tabularx}{\linewidth}{XX}
        \thickhline
        \includegraphics[width=\linewidth,valign=t]{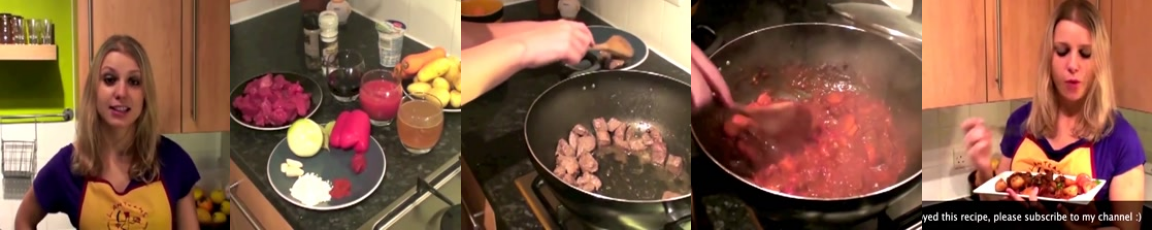} &
        \includegraphics[width=\linewidth,valign=t]{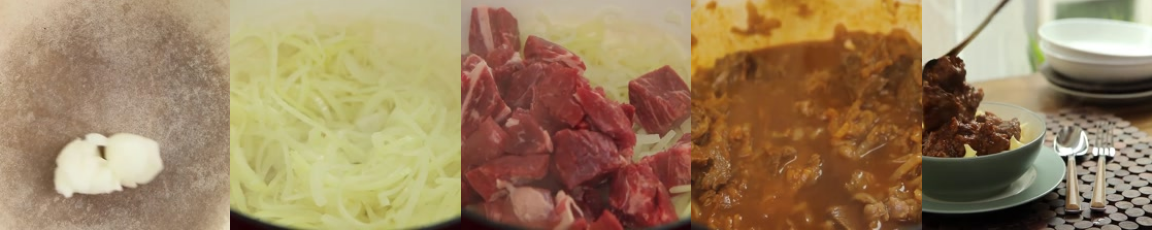} \\

        \begin{minipage}[t]{0.48\textwidth}
            \textbf{MART}: Cook the bacon in a pan. Add chopped onions to the pan. Add chopped carrots. Add chopped
            tomatoes to the pan. Add the chicken to the pan. \\

            \textbf{COOT (Ours)}: Fry the beef in a pan. Add onion and carrot to the pan. Add the chicken to the pan.
            Add the tomatoes and stir. Add the potatoes to the pan. \\

            \textbf{GT}: Brown 400gm of sliced beef on a hot pan. Fry onions until golden then add garlic carrots and
            red pepper fry for 5 mins. Now add the beef 2 tbsp of flour 1 tsp of paprika 1 tbsp of tomato puree 2 bay
            leaves and 300ml beef stock. Add 200 gram canned tomato 100ml red wine sour cream and mix well let it
            simmer for 1 5 hour. Now add 400gm of baby potato and mix it let it cook for 30 more min. \\

        \end{minipage}
        &

        \begin{minipage}[t]{0.48\textwidth}
            \textbf{MART}: Add flour to a bowl and whisk. Cut the chicken into pieces. Coat the chicken in flour.
            Coat the chicken in flour egg and breadcrumbs. Fry the chicken in a pan. Drizzle the sauce on top of the
            bread. Add sauce to the pizza. Bake the dish in the oven. \\

            \textbf{COOT (Ours)}: Mix parmesan cheese black pepper and garlic powder. Cover the chicken in the bag.
            Coat the chicken in the flour. Coat the chicken in the egg and coat with flour. Place the chicken in a
            pan and fry it on a pan. Pour sauce on top of the chicken and top with mozzarella cheese. Sprinkle
            parmesan cheese on top. Bake the chicken in the oven. \\

            \textbf{GT}: Mix bread crumbs and parmesan cheese. Pound the chicken. Rub salt and pepper onto the
            chicken. Rub flour onto the chicken dip it in egg and coat with breadcrumbs. Fry the chicken in a pan.
            Spread sauce over the chicken. Top the chicken with mozzarella cheese. Bake the chicken in the oven. \\

        \end{minipage}
        \\

        \includegraphics[width=\linewidth,valign=t]{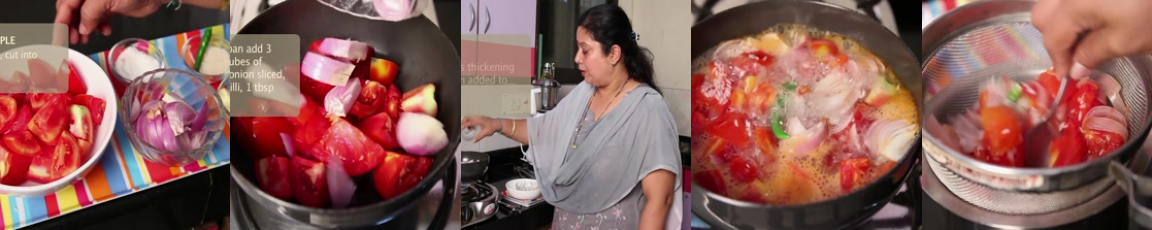} &
        \includegraphics[width=\linewidth,valign=t]{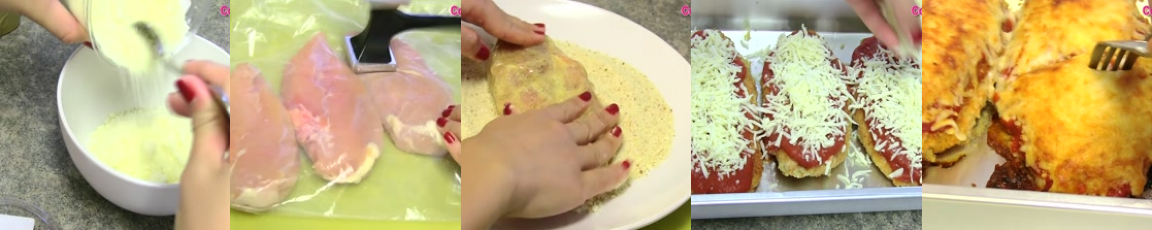} \\

        \begin{minipage}[t]{0.48\textwidth}
            \textbf{MART}: Add tomatoes and beef to a pot. Add water to the pan. Add tomato puree and salt. Add the
            beef and parsley to the soup. Add the beef to the pot. Add water to the soup and let it simmer. Add the
            soup to the soup. \\

            \textbf{COOT (Ours)}: Add the tomatoes and onions to a food processor and blend them. Add the tomatoes
            and a bay leaf to the pot. Add the tomatoes and simmer. Remove the tomatoes from the pot and let it cook.
            Remove the tomatoes from the pot and let it cook. Strain the soup to a boil and let it boil. Turn on the
            heat and heat to a boil. \\

            \textbf{GT}: Add tomato onion green chili and rice to a pan. Add water to the pan. Boil the ingredients
            and then turn down the heat. Strain the ingredients. Blend the ingredients. Add the water to the mixture
            and strain. Boil the soup. \\

        \end{minipage}
        &

        \begin{minipage}[t]{0.48\textwidth}
            \textbf{MART}: Add flour to a bowl and whisk. Cut the chicken into pieces. Coat the chicken in flour.
            Coat the chicken in flour egg and breadcrumbs. Fry the chicken in a pan. Drizzle the sauce on top of the
            bread. Add sauce to the pizza. Bake the dish in the oven. \\

            \textbf{COOT (Ours)}: Mix parmesan cheese black pepper and garlic powder. Cover the chicken in the bag.
            Coat the chicken in the flour. Coat the chicken in the egg and coat with flour. Place the chicken in a
            pan and fry it on a pan. Pour sauce on top of the chicken and top with mozzarella cheese. Sprinkle
            parmesan cheese on top. Bake the chicken in the oven. \\

            \textbf{GT}: Mix bread crumbs and parmesan cheese. Pound the chicken. Rub salt and pepper onto the
            chicken. Rub flour onto the chicken dip it in egg and coat with breadcrumbs. Fry the chicken in a pan.
            Spread sauce over the chicken. Top the chicken with mozzarella cheese. Bake the chicken in the oven. \\

        \end{minipage}
        \\
\thickhline
    \end{tabularx}

    \label{tab:cap_yc2_random}
\end{table}

\begin{table}[t]
    \caption{
    \textbf{Random Captioning samples on ActivityNet (ae-val split).}
    }
    \centering
    \small
    \setlength\tabcolsep{\tabcolsepqual}
    \renewcommand*{\arraystretch}{\arraystretchqual}

    \begin{tabularx}{\linewidth}{XX}
        \thickhline
        \includegraphics[width=\linewidth,valign=t]{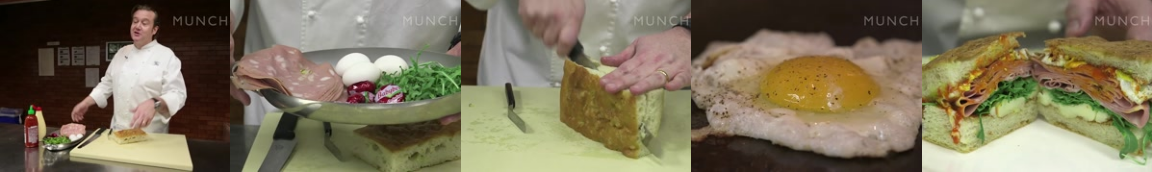} &
        \includegraphics[width=\linewidth,valign=t]{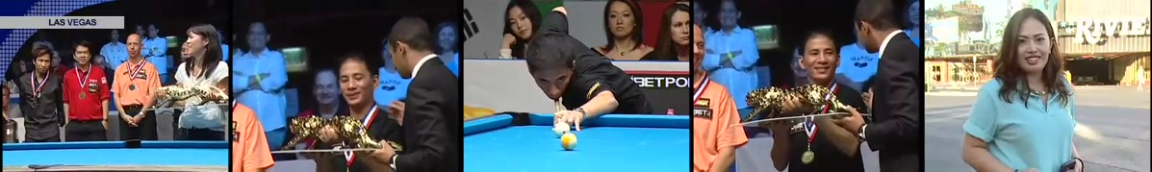} \\

        \begin{minipage}[t]{0.48\textwidth}
            \textbf{MART}: A man is seen speaking to the camera and leads into him holding up various objects and
            presenting them to. He then cuts the knife and cuts the sandwich while still speaking to the camera. He
            then puts the sandwich into the pan and cuts it in half. He then puts the sandwich into the sandwich and
            puts it in the end. \\

            \textbf{COOT (Ours)}: A chef demonstrates how to make a sandwich using bread , then he puts a knife in a
            kitchen and. Then , the man puts the bread on a bread and cuts it in half. After , the man puts the
            sandwich in the bread and put it in a plate. Next , the man cuts the bread and put on top of the sandwich
            . \\

            \textbf{GT}: A man shows ingredients for a mortadella sandwich. The man cuts the bred in four pieces and
            puts mustard and then brown on the stove. Then, the man fries an egg and puts it on the bread as well the
            mortadella, green leaves, cheese and ketchup. After, the man cuts the sandwich in two and eat one. \\

        \end{minipage}
        &

        \begin{minipage}[t]{0.48\textwidth}
            \textbf{MART}: A person is seen sitting in front of a large pile of grass and holding a stick. The person
            then puts the tire on the machine and begins putting the tire on. \\

            \textbf{COOT (Ours)}: A person is seen using a tool on a machine and piecing together with the camera.
            The man continues to use the machine on the machine and ends by taking out more out of the machine. \\

            \textbf{GT}: A person is seen walking in with a tire on a plank and painting the tire. The person then un
            does the tire and places the rubber tightly around the side. \\

        \end{minipage}
        \\

        \includegraphics[width=\linewidth,valign=t]{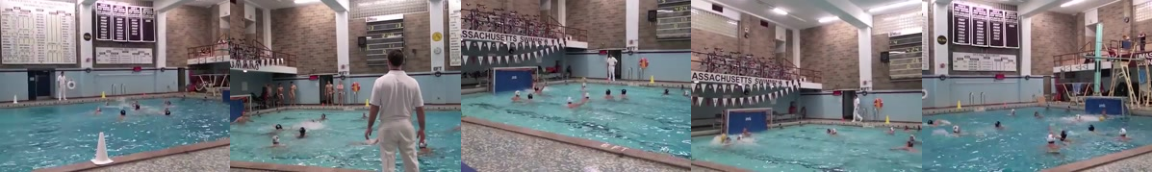} &
        \includegraphics[width=\linewidth,valign=t]{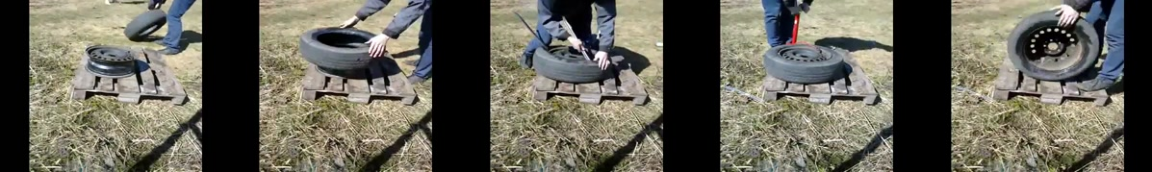} \\

        \begin{minipage}[t]{0.48\textwidth}
            \textbf{MART}: A small group of people are seen swimming around a pool throwing a ball around to one
            another. The people continue playing with one another and end by throwing the ball back and fourth. \\

            \textbf{COOT (Ours)}: A large group of people are seen swimming around a pool throwing a ball around to
            one another. The people continue playing with one another and ends with a large group of people watching
            on the sides. \\

            \textbf{GT}: A video of water polo is shown in the gym. A few people watch and the ball goes back and
            forth. \\

        \end{minipage}
        &

        \begin{minipage}[t]{0.48\textwidth}
            \textbf{MART}: A person is seen sitting in front of a large pile of grass and holding a stick. The person
            then puts the tire on the machine and begins putting the tire on. \\

            \textbf{COOT (Ours)}: A person is seen using a tool on a machine and piecing together with the camera.
            The man continues to use the machine on the machine and ends by taking out more out of the machine. \\

            \textbf{GT}: A person is seen walking in with a tire on a plank and painting the tire. The person then
            undoes the tire and places the rubber tightly around the side. \\

        \end{minipage}
        \\
\thickhline
    \end{tabularx}

    \label{tab:cap_anetval_random}
\end{table}

\begin{table}[t]
    \caption{
        \textbf{Random Captioning samples on ActivityNet (ae-test split).}
        }
    \centering
    \small
    \setlength\tabcolsep{\tabcolsepqual}
    \renewcommand*{\arraystretch}{\arraystretchqual}

\begin{tabularx}{\linewidth}{XX}
\thickhline
\includegraphics[width=\linewidth,valign=t]{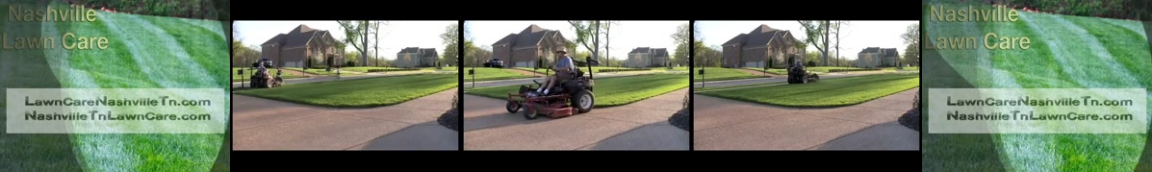}&
\includegraphics[width=\linewidth,valign=t]{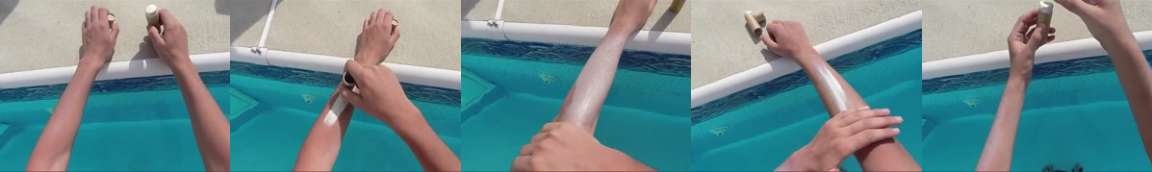}\\

\begin{minipage}[t]{0.48\textwidth}
\textbf{MART}: A woman stands on front a house talking. The woman drives the lawn mower with a mower. The woman drives the lawn mower. The woman pushes the lawn mower along the grass. The woman talks to the camera. \\

\textbf{COOT (Ours)}: We see the title on the field , white and white text. We then see a man mowing his lawn. The man stops and talks to the camera. The man stops and turns around. We then see the grass again. \\

\textbf{GT}: The video begins with a picture of a lawn along with a company name and website. The video cuts to a man riding a lawnmower, cutting the grass in a nice neighborhood. When he begins, some kids are playing in the road. At one point, a car passes by. The video ends with the picture of the lawn showing the company name and website. \\

\end{minipage}
&

\begin{minipage}[t]{0.48\textwidth}
\textbf{MART}: A group of women are dancing on a stage. They are dancing together in a room. They are dancing together. \\

\textbf{COOT (Ours)}: A large group of girls are seen standing together followed by a woman dancing and performing a dance routine. The woman continues speaking to the camera while more people are seen dancing around and leads into a group of. The group continues dancing with one another and ends with a woman speaking to the camera. \\

\textbf{GT}: Several girls are in a classroom dancing and doing ballet. The instructor then comes to talk briefly before continuing on coaching the girls. After,the exercises continue and the girls do leaps and jumps in the room before the outside of the dance studio is shown. \\

\end{minipage}
\\

\includegraphics[width=\linewidth,valign=t]{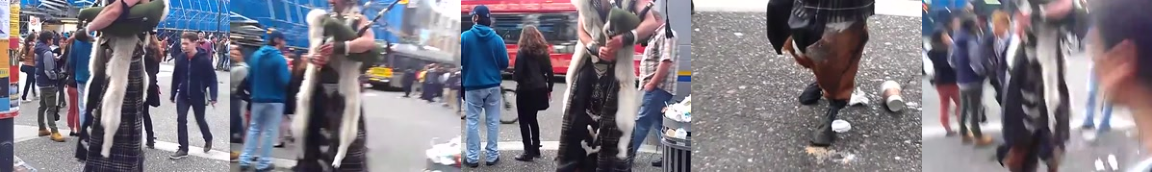}&
\includegraphics[width=\linewidth,valign=t]{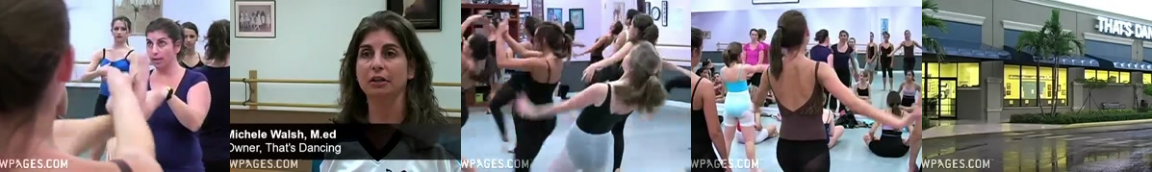}\\

\begin{minipage}[t]{0.48\textwidth}
\textbf{MART}: People are gathered around a street watching. They are holding flags in their hands. A man in a white shirt is standing next to a fence. \\

\textbf{COOT (Ours)}: A man plays bagpipes while people watch on the sidewalk. A person in a black shirt plays the bagpipes. A person in a white shirt walks past the person. \\

\textbf{GT}: A man on stilts is playing the bag pipes on a street. A bus passes on the street behind the man. A street sign on a pole is shown. \\

\end{minipage}
&

\begin{minipage}[t]{0.48\textwidth}
\textbf{MART}: A group of women are dancing on a stage. They are dancing together in a room. They are dancing together. \\

\textbf{COOT (Ours)}: A large group of girls are seen standing together followed by a woman dancing and performing a dance routine. The woman continues speaking to the camera while more people are seen dancing around and leads into a group of. The group continues dancing with one another and ends with a woman speaking to the camera. \\

\textbf{GT}: Several girls are in a classroom dancing and doing ballet. The instructor then comes to talk briefly before continuing on coaching the girls. After,the exercises continue and the girls do leaps and jumps in the room before the outside of the dance studio is shown. \\

\end{minipage}
\\
\thickhline
\end{tabularx}

    \label{tab:cap_anettest_random}
\end{table}

\begin{figure}
    \centering
    \includegraphics[width=1\linewidth]{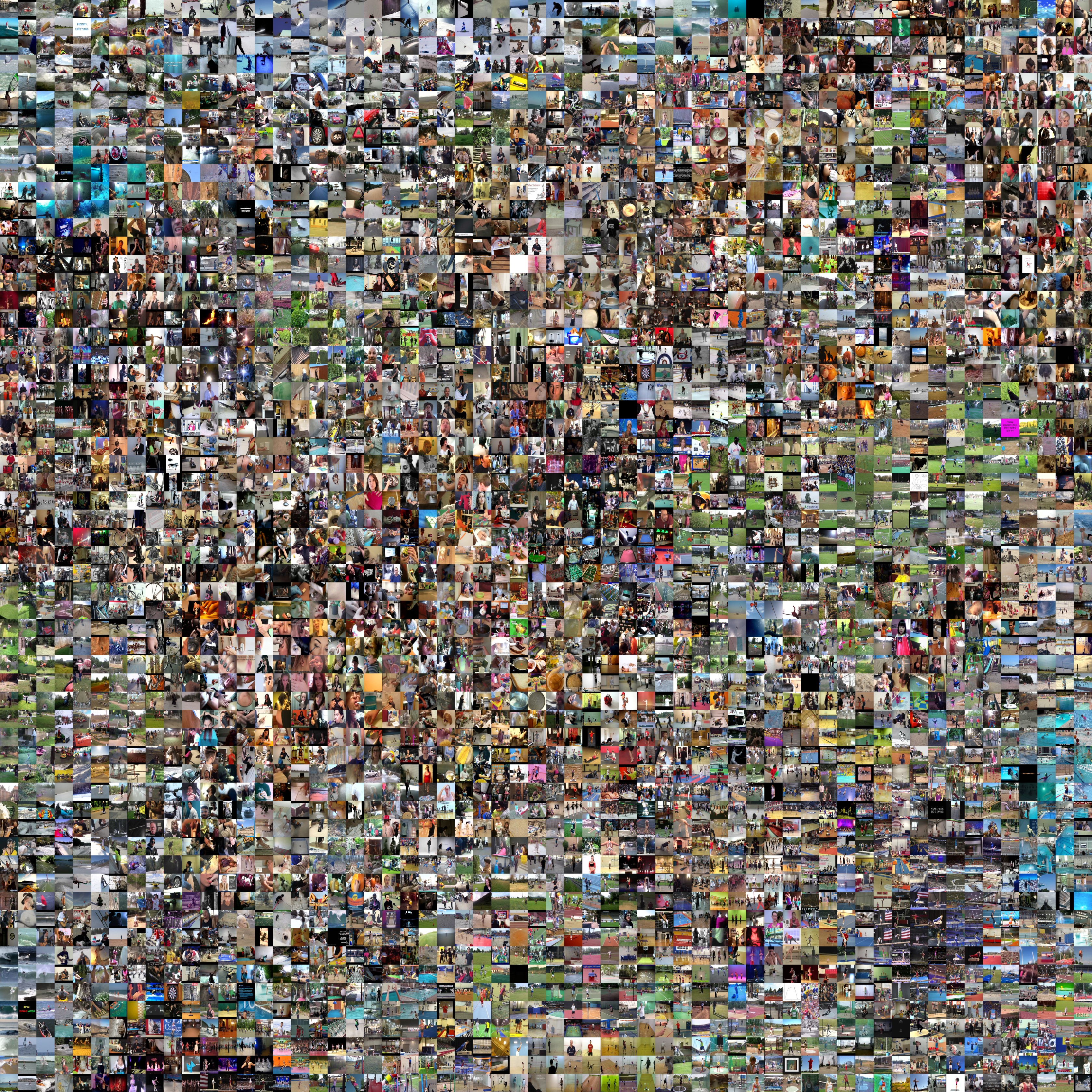}
    \caption{
        \textbf{Visualization of the video embedding space with t-SNE on ActivityNet-Captions.}
        We apply t-SNE to reduce the video embedding space to 2 dimensions and visualize videos by one sample frame.
    }
    \label{fig:tsne}
\end{figure}

\narrowparagraph{ActivityNet-Captions.}
To further check whether our COOT model can learn the semantic alignment of video and text well,
we provide qualitative examples for the retrieval task on the ActivityNet-caption dataset (val1 split, 4917
video-paragraph pairs).
Note that any spelling errors in the dataset are not corrected. As shown in Table~\ref{tab:retr_anet_p2v} and
Table~\ref{tab:retr_anet_v2p},
the model learns to semantically align the video and paragraph embeddings. Even for imperfect rankings, the model
retrieves semantically similar items.

\narrowparagraph{YouCook2.}
We also present a set of qualitative clip-to-sentence and sentence-to-clip retrieval examples for the YouCook2 dataset
(val split, 3492 clip-sentence pairs, 457 video-paragraph pairs).
Table~\ref{tab:retr_yk_p2v} and Table~\ref{tab:retr_yk_v2p} show several examples where we can reasonably retrieve
similar semantics,
even when the wrong object is recognized (Table~\ref{tab:retr_yk_p2v}-Right).

\narrowparagraph{t-SNE Visualization of Embeddings.}
We project the video embeddings of Activitynet dataset to 2D space using t-SNE~\cite{maaten2008visualizing} and
visualize each point with a sample frame from the video.
As shown in Figure~\ref{fig:tsne}, the embeddings are clustered semantically around activities and
videos with similar content are in close neighborhood.

\subsection{Captioning Results}\label{subsec:capt}
To expand upon the qualitative captioning results, we provide evaluation on samples that are not cherry-picked for
Youcook2 (val split) and ActivityNet (ae-val and ae-test split) in Tables ~\ref{tab:cap_yc2_random},
~\ref{tab:cap_anetval_random}, ~\ref{tab:cap_anettest_random}.
\end{document}